\documentclass[a4paper,10pt]{article}

\usepackage{graphicx}
\usepackage{amssymb}
\usepackage{url}
\usepackage[colorlinks = true,
            linkcolor = magenta,
            urlcolor  = blue,
            citecolor = green,
            anchorcolor = blue]{hyperref}

\usepackage[normalem]{ ulem }
\usepackage{soul}

\usepackage{microtype}
\usepackage{graphicx}
\usepackage{subcaption}
\usepackage{enumitem}
\usepackage{booktabs}
\usepackage{verbatimbox}
\usepackage{cuted}
\usepackage{authblk}

\newcommand{\figref}[1]{Figure \ref{#1}}
\newcommand{\bs}[1]{\boldsymbol{#1}}

\usepackage{amsmath,amsfonts,bm,bbm}

\def\figref#1{figure~\ref{#1}}

\def\secref#1{section~\ref{#1}}

\def\1{\bm{1}}

\def\rmtheta{{\mathbf{\Theta}}}

\def\vtheta{{\bm{\theta}}}
\def\va{{\bm{a}}}

\def\vx{{\bm{x}}}

\DeclareMathAlphabet{\mathsfit}{\encodingdefault}{\sfdefault}{m}{sl}
\SetMathAlphabet{\mathsfit}{bold}{\encodingdefault}{\sfdefault}{bx}{n}

\def\gO{{\mathcal{O}}}

\def\sR{{\mathbb{R}}}

\DeclareMathOperator*{\E}{\mathbb{E}}

\newcommand\at[2]{\left.#1\right|_{#2}}

\usepackage[top=2.5cm, bottom=2.5cm, left=2cm, right=2cm]{geometry}

\begin{document}

\title{Double Trouble in Double Descent: \\Bias and Variance(s) in the Lazy Regime}

\author{St\'ephane d'Ascoli$^*$, Maria Refinetti$^*$, Giulio Biroli and Florent Krzakala}

\date{Laboratoire de Physique de l'Ecole Normale Sup\'erieure, Universit\'e PSL, \\CNRS, Sorbonne
Universit\'e, F-75005 Paris, France\\ \footnotesize{$*$ Equal contribution.}}

\setcounter{Maxaffil}{0}
\renewcommand\Affilfont{\itshape\small}

\maketitle

\begin{abstract}
Deep neural networks can achieve remarkable generalization performances while interpolating the training data perfectly. Rather than the U-curve emblematic of the bias-variance trade-off, their test error often follows a ``double descent"--- a mark of the beneficial role of overparametrization. In this work, we develop a quantitative theory for this phenomenon in the so-called lazy learning regime of neural networks, by considering the problem of learning a high-dimensional function with random features regression. We obtain a precise asymptotic expression for the bias-variance decomposition of the test error, and show that the bias displays a phase transition at the interpolation threshold, beyond which it remains constant. We disentangle the variances stemming from the sampling of the dataset, from the additive noise corrupting the labels, and from the initialization of the weights.  Following up on Geiger et al. \cite{geiger2019scaling}, we first show that the latter two contributions are the crux of the double descent: they lead to the overfitting peak at the interpolation threshold and to the decay of the test error upon overparametrization. We then quantify how they are suppressed by ensembling the outputs of $K$ independently initialized estimators. When $K$ is sent to infinity, the test error remains constant beyond the interpolation threshold. We further compare the effects of overparametrizing, ensembling and regularizing. Finally, we present numerical experiments on classic deep learning setups to show that our results hold qualitatively in realistic lazy learning scenarios.  
\end{abstract}

\setcounter{tocdepth}{2}

Correspondence: \href{mailto:stephane.dascoli@ens.fr}{stephane.dascoli@ens.fr}, \href{mailto:maria.refinetti@ens.fr}{maria.refinetti@ens.fr}.

\section{Introduction}
\label{sec:introduction}
Deep neural networks have achieved breakthroughs in a plethora of contexts, such as image classification~\cite{krizhevsky2012imagenet, lecun2015deep}, speech recognition~\cite{hinton2012deep}, and automatic translation~\cite{sutskever2014sequence}.
Yet, theory lags far behind practice, and the key reasons underpinning the success of deep learning remain to be clarified. 

One of the main puzzles is to understand the excellent generalization performance of heavily overparametrized deep neural networks able to fit random labels~\cite{zhang2016understanding}.
Such \emph{interpolating} estimators---that can reach zero training error--- have attracted a growing amount of theoretical attention in the last few years, see e.g.~\cite{advani2017high,belkin2018reconciling,neal2018modern,hastie2019surprises,mei2019generalization}. Indeed, classical learning theory suggests that generalization should first improve then worsen when increasing model complexity, following a U-shape curve characteristic of the bias-variance trade-off. Instead, deep neural networks~\cite{neyshabur2014search,spigler2018jamming,nakkiran2019deep} as well as other machine learning models~\cite{belkin2018reconciling}, follow a different curve, coined \emph{double descent}.

This curve displays two regimes :  
the \emph{classical} U-curve is superseded at high complexity by a \emph{modern} interpolating regime where the test error decreases monotonically with overparametrization \cite{breiman1995reflections}. 
Between these two regimes, i.e. at the \emph{interpolation threshold} where training error vanishes, a \emph{peak} occurs in absence of regularization, sometimes called the \emph{jamming} peak {\color{blue} \cite{spigler2018jamming}} due to similarities with a well-studied phenomenon in the Statistical Physics litterature~\cite{opper1996statistical,liu1998jamming,engel2001statistical,franz2016simplest,krzakala2007landscape,zdeborova2007phase}. The reasons behind the performance of deep neural networks in the overparametrized regime are still poorly understood, even though some mechanisms are known to play an important role, such as the implicit regularization of stochastic gradient descent which allows to converge to the minimum norm solution, and the convergence to mean-field limits \cite{rosset2004margin,advani2017high,chizat2018note,arora2019exact,mei2019mean}.

Here we present a detailed investigation of the double descent phenomenon, and its theoretical explanation in terms of bias and variance in the so-called \emph{lazy} regime \cite{chizat2018note}. 
This theoretically appealing scenario, where the weights stay close to their initial value during training, is called \emph{lazy learning} as opposed to \emph{feature learning} where the weights change enough to learn relevant features~\cite{chizat2018note,woodworth2019kernel,geiger2019disentangling}. Although replacing learnt features by random features may appear as a crude simplification, empirical results show that the loss in performance can be rather small in some cases~\cite{arora2019harnessing,arora2019exact}. A burst of recent papers showed that in this regime, neural networks behave like kernel methods \cite{daniely2016toward,lee2017deep,jacot2018neural} or equivalently random projection methods \cite{rahimi2008random,chizat2018note,arora2019exact}. This mapping makes the training analytically tractable, allowing, for example, to prove convergence to zero error solutions in overparametrized settings.

Optimization plays an important role in neural networks, and in particular for the double descent phenomenon, by inducing implicit regularization~\cite{neyshabur2017geometry} and fluctuations of the learnt estimator~\cite{geiger2019scaling}. Disentangling the variance stemming from the randomness of the optimization process from that the variance due to the randomness of the dataset is a crucial step towards a unified picture, as suggested in~\cite{neal2018modern}. In this paper, we address this issue and attempt to reconcile the behavior of bias and variance with the double descent phenomenon by providing a precise and quantitative theory in the \emph{lazy} regime.

\paragraph{Contributions}
We focus on an analytically solvable model of random features (RF), introduced by \cite{rahimi2008random}, that can be viewed either as a randomized approximation to kernel ridge regression, or as a two-layer neural network whose first layer contains fixed random weights. The latter provides a simple model for lazy learning. Indeed, suppose that a neural network learns a function $f_{\theta}({\bf x})$ that relates labels (or responses) $y$ to inputs ${\bf x}$ via a set of weights ${\theta}$.
The lazy regime is defined as the setting where the model can be linearized around the initial conditions $\theta_0$. Assuming that the initialization is such that $f_{\theta_0} \approx 0$\footnote{One can alteratively define the estimator as $f_{\theta}-f_{\theta_0}$ \cite{chizat2018note}.}, one obtains:
\begin{equation}
f_{\theta}({\bf x}) \approx \left.\nabla_{\theta}f_{\theta}({\bf x})\right|_{\theta=\theta_0} \cdot \bf (\theta-\theta_0) \,.
\end{equation}
In other words, the lazy regime corresponds to a linear fitting problem with a random feature vector $\left.\nabla_{\theta}f_{\theta}({\bf x})\right|_{\theta=\theta_0}$. In this setting our contributions are:
\begin{itemize}
\item  We demonstrate how to disentangle quantitatively the contributions to the test error of the bias and the various sources of variance of the estimator, stemming from the sampling of the dataset, from the additive noise corrupting the labels, and from the initialization of the random feature vectors.
\item We give a sharp asymptotic formula for the effect of \emph{ensembling} (averaging the predictions of indepently initialized estimators) on these various terms. We show in particular how the over-fitting peak at the interpolation threshold can be attenuated by ensembling, as observed in real neural networks~\cite{geiger2019scaling}. We also compare the effect of ensembling, overparametrizing and optimally regularizing. 
\item Several conclusions stem from the above analysis. First, the over-fitting near the interpolation threshold is entirely due to the variances due to the additive noise in the ground truth and the initialization of the random features. Second, the data sampling variance and the bias both display a phase transition at the interpolation threshold, and remain constant in the overparametrized regime. Hence, the benefit of ensembling and overparametrization beyond the interpolation threshold is solely due to a reduction of the noise and initialization variances. 
\end{itemize}

Finally, we present numerical results on a classic deep learning scenario in the 
lazy learning regime to show that our findings, obtained for simple random features and i.i.d. data, are relevant to realistic setups involving correlated random features and realistic data. 

The analytical results we present are obtained using a heuristic method from Statistical Physics called the Replica Method~\cite{mezard1987spin}, which despite being non-rigorous has shown its remarkable efficacy in many machine learning problems~\cite{seung1992statistical,engel2001statistical,advani2013statistical,zdeborova2016statistical} and random matrix topics, see e.g. ~\cite{livan2018introduction, tarquini2016level,aggarwal2018goe}. While it is an open problem to provide a rigorous proof of our computations, we check through numerical simulations that our asymptotic predictions are extremely accurate at moderately small sizes.

\paragraph{Related work} 

Our work crucially builds on two recent contributions by Geiger {\it et al.}~\cite{geiger2019scaling}, and Mei and Montanari~\cite{mei2019generalization}. The authors of~\cite{geiger2019scaling} carried out a series of experiments in order to shed light on the generalization properties of neural networks. The current work is  inspired by their observations and  scaling theory about the role of the variance due to the random initialization of the weights in the double-descent curve. They argued that the decrease of the test error in the limit of very wide networks is due to this source of variance, which vanishes inversely proportional to the width of the network. They then used ensembling to empirically support these findings in more realistic situations. Another related work is \cite{neal2018modern}, which empirically disentangles the various sources of variance in the process of training deep neural networks. 
 
On the analytical side, our paper builds on the results of \cite{mei2019generalization}, which provide a precise expression of the test error of the RF model in the high-dimensional limit where the number of random features, the dimension of the input data and the number of data points are sent to infinity with their relative ratios fixed. The double descent was also studied analytically for various types of linear models, both for regression~\cite{advani2017high,hastie2019surprises,nakkiran2019more,belkin2019two} and classification~\cite{deng2019model,kini2020analytic,replica2020}. An example of a practical method that uses ensembling in kernel methods is detailed in \cite{drucker1994boosting}. Note that this work performs an average over the sampling of the random feature vectors in contrast to \cite{zhang2013divide} where the average is taken over the sampling of the data set.

\paragraph{Reproducibility} The codes necessary to reproduce the results presented in this paper and obtain new ones are given at~ \href{https://github.com/mariaref/Random_Features.git
}{https://github.com/mariaref/Random\textunderscore Features.git}.

\section{Model}
\label{sec:model}
\begin{figure}[htb]
    \centering
		\includegraphics[width=0.5\linewidth]{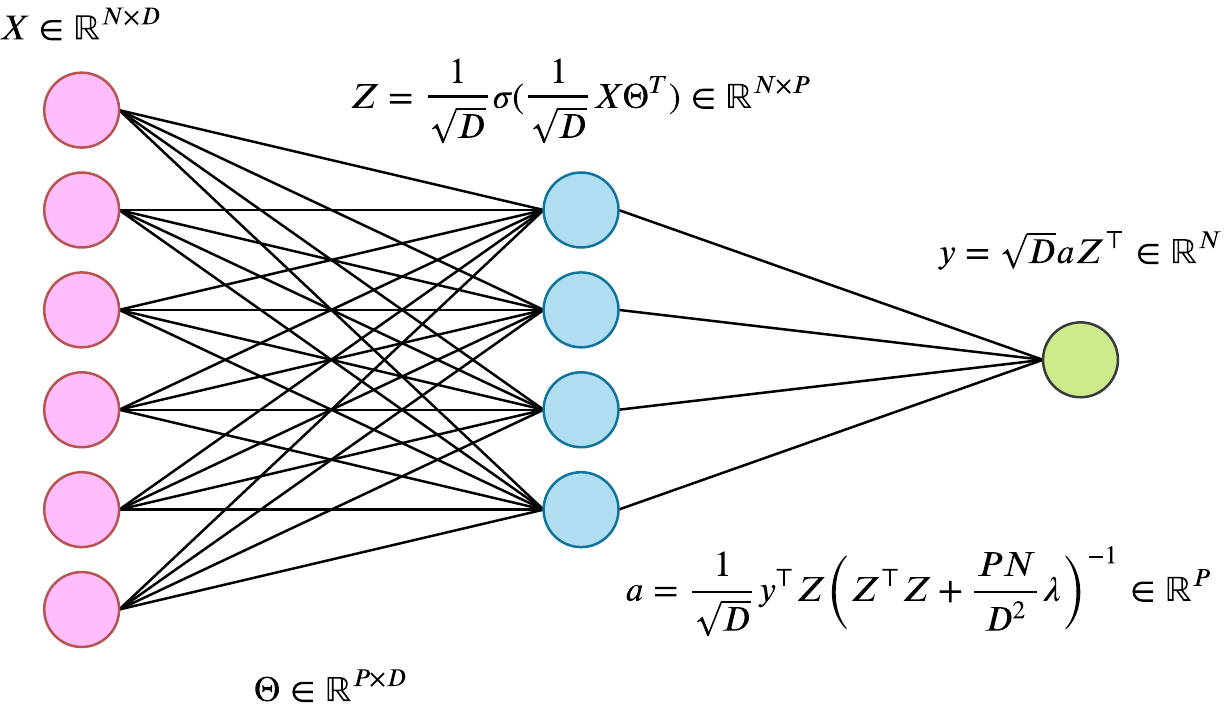}
		\caption{Illustration of a Random Feature (RF) network. The first layer weights are fixed and initialized as i.i.d. centered Gaussian variables of unit variance. The second layer weights are trained via ridge regression.}
		    \label{fig:rf-single}
\end{figure}

This work is centered around the RF model first introduced in \cite{rahimi2008random}. Although simpler settings such as linear regression display the double descent phenomenology~\cite{hastie2019surprises}, this  model is more appealing in several ways. First, the presence of two layers allows to freely disentangle the dimensionality of the input data from the number of parameters of the model. Second, it closely relates to the lazy regime of neural networks, as described above. Third, and most importantly for our specific study, the randomness of the fixed first layer weights mimics the randomness due to weight initialization in neural networks.

The RF model can be seen as a two-layer neural network whose first layer contains fixed random weights\footnote{Note the closeness between the RF model and the "\emph{hidden manifold model}" introduced in \cite{goldt2019modelling}. The task studied here can be seen as a linear regression task on a structured data set $\bs Z\in\mathbb{R}^P$, obtained by projecting the original \emph{latent} features $\bs X\in\mathbb{R}^D$. The difference here is that the dimension of the latent space, denoted as $D$ here, is sent to infinity together with the dimension of the ambient space.} (see \figref{fig:rf-single}):
    \begin{equation}
        \hat f(\vx)=\sum_{i=1}^P\va_i \sigma\left(\frac{\langle\vtheta_i, \vx\rangle}{\sqrt{D}}\right).
\end{equation}
In the above, $\vtheta_i$ is the $i^{th}$ \emph{random feature vector}, i.e the $i^{th}$ column of the random feature matrix $\rmtheta\in \mathbb{R}^{P \times D}$ whose elements are drawn i.i.d from $\mathcal{N}(0,1)$. $\sigma$ is a pointwise activation function, which we will take to be $ReLU:x\mapsto \max (0,x)$.

The training data is collected in a matrix $\bs X \in \mathbb{R}^{N\times D}$ whose elements are drawn i.i.d from $\mathcal{N}(0,1)$. We assume that the labels are given by a linear ground truth corrupted by some additive Gaussian noise:
\begin{align}
    &y_\mu =  \langle\boldsymbol{\beta},\bs X_\mu\rangle+\epsilon_\mu, \quad ||\boldsymbol{\beta}|| = F, \quad \epsilon_\mu\sim\mathcal{N}(0,\tau),\\
    &\mathrm{SNR} = F/\tau.\notag
\end{align}
The generalization to non-linear functions can also be performed as in~\cite{mei2019generalization}.

The second layer weights, i.e the elements of $\va$, are calculated by the means of ridge regression:
\begin{equation}
\begin{split}
    {\cal L}_{\mathrm{RF}}(\boldsymbol{a}) &\equiv \frac{1}{N}\sum_{\mu=1}^N \left(\!y_\mu\!-\!\sum_{i=1}^P\va_i \sigma\!\left(\frac{\langle\vtheta_i, \bs X_\mu\rangle}{\sqrt{D}}\right)\!\!\right)^{2}\!\!\!\!+\frac{P \lambda}{D}\|\boldsymbol{a}\|_{2}^{2}, \\
    \hat{\boldsymbol{a}} &\equiv \underset{\boldsymbol{a} \in \mathbb{R}^{P}}{\arg \min } ~ {\cal L}_{\mathrm{RF}}(\boldsymbol{a}).
\end{split}
\label{eq:estimator}
\end{equation}
Note that as $P\to \infty$, this is equivalent to kernel ridge regression with respect to the following kernel: 
\begin{equation*}
\begin{split}
     K\left(\boldsymbol{x}, \boldsymbol{x}^{\prime}\right)=\E_{\boldsymbol{\theta} \sim \mathcal{P}}\left[\sigma(\langle\boldsymbol{x}, \boldsymbol{\theta}\rangle / \sqrt{D}) \sigma\left(\left\langle\boldsymbol{x}^{\prime}, \boldsymbol{\theta}\right\rangle / \sqrt{D}\right)\right],
\end{split}
\end{equation*}
where $\mathcal{P} = \operatorname{Unif}\left(\mathbb{S}^{D-1}(\sqrt{D})\right)$.

The key quantity of interest is the \emph{test error} of this model, defined as the mean square error evaluated on a fresh sample $\vx\sim \mathcal{N}(0,1)$ corrupted by a new noise $\tilde \epsilon$:
\begin{align}
    &\mathcal{R}_{\mathrm{RF}} =\E_x \left[ \left( \langle \bs \beta, \bs x\rangle + \tilde \epsilon - \hat f(\bs x) \right)^2 \right], \quad \tilde \epsilon \sim \mathcal{N}(0,\tilde \tau).
    \label{eq:gen-error-a}
\end{align}

\section{Analytical results}
\label{sec:analytics}
In this section, we present our main result, which is an analytical expression for the terms appearing in the decomposition of the test error.

\paragraph{Decomposition of the test error}

The test error can be decomposed into its bias and variance components:
\begin{align}
    \E_{\bs \Theta, \bs X, \boldsymbol{\varepsilon}}\left[ \mathcal{R}_{\mathrm{RF}}\right]\!=\!
    \mathcal{E}_{\rm Noise}\!+\!\mathcal{E}_{\rm Init}\!+\!\mathcal{E}_{\rm Samp}\!+\!\mathcal{E}_{\rm Bias}\!+\! \tilde\tau^2.
    \label{eq:decomposition}
\end{align}
The first three terms contribute to the variance, the fourth is the bias, and the final term $\tilde\tau^2$ is simply the Bayes error. It does not play any role and will be set to zero in the rest of the paper: the only reason it was included is to avoid confusion with $\mathcal{E}_{\rm Noise}$ defined below.

\textbf{Noise variance}: The first term is the variance associated with the additive noise corrupting the labels of the dataset which is learnt, $\bs \varepsilon$:
\begin{equation}
\begin{split}
       \mathcal{E}_{\rm Noise}&=\E_{\bs{x},\bs{X},\bs \Theta } \left[\E_{\bs{\varepsilon}}\left[\hat{f}(\bs{x})^2\right]-\left(\E_{\bs{\varepsilon}}\left[\hat{f}(\bs{x})\right]\right)^2\right].
\end{split}
    \label{eq:var_noise}
\end{equation}

\textbf{Initialization variance}: The second term encodes the fluctuations stemming from the random initialization of the random feature vectors, $\bs \Theta$:
\begin{align}
     \mathcal{E}_{\rm Init}&=\E_{\bs{x},\bs{X}} \left[\E_{\bs{\Theta}}
    \left[\E_{\bs{\varepsilon}}\left[\hat{f}(\bs{x})\right]^2\right] -\E_{\bs{\Theta},\bs{\varepsilon}}\left[\hat{f}(\bs{x})\right]^2\right].
    \label{eq:var_init}
\end{align}

\textbf{Sampling variance}: The third term measures the fluctuations due to the sampling of the training data, $\bs X$:
\begin{equation}
    \mathcal{E}_{\rm Samp}=\E_{\bs{x}} \left[\E_{\bs{X}}\left[\E_{\bs{\Theta},\bs{\varepsilon} }\left[\hat{f}(\bs{x})\right]^2\right]
    -\!\!\!\E_{\bs{X},\bs{\Theta}, \bs{\varepsilon} }\left[\hat{f}(\bs{x})\right]^2\right].
\label{eq:var_samp}
\end{equation}

\textbf{Bias}: Finally, the fourth term is the bias, i.e. the error that remains once all the sources of variance have been averaged out. It can be understood as the approximation error of our model and takes the form:
\begin{align}
    \mathcal{E}_{\rm Bias}&=\E_{\bs{x}}\left[\left(\langle\bs\beta, \bs{x}\rangle-\E_{\bs{X},\bs{\Theta},\bs{\varepsilon}}\left[\hat{f}(\bs{x})\right]\right)^2\right].
    \label{eq:bias}
\end{align}
Note that since we are performing deterministic ridge regression, the noise induced by SGD, which can play an important role outside the lazy regime for deep neural networks, cannot be captured. 

\paragraph{Main result} Consider the high-dimensional limit where the input dimension $D$, the hidden layer dimension $P$ (which is equal to the number of parameter in our model) and the number of training points $N$ go to infinity with their ratios fixed:
    \begin{equation}
        N,P,D\to\infty, \quad \frac{P}{D}=\gO(1),\quad \frac{N}{D}=\gO(1).
    \end{equation}

We obtain the following result:
\begin{align}
    \E_{\bs x,\bs\varepsilon,\bs \Theta,\bs X} \left[\langle\bs\beta,\bs x\rangle\hat{f}(\bs x)\right]=&F^2 \Psi_1,\label{eq:1}\\
    \E_{\bs x,\bs \Theta,\bs X}\left[\E_{\bs\varepsilon}\left[\hat{f}(\bs x)\right]^2\right]=&F^2\Psi_2^v,\label{eq:2}\\
    \E_{\bs x,\bs \Theta,\bs X} \left[\E_{\bs\varepsilon}\left[\hat{f}(\bs x)^2\right]\!-\!\E_{\bs\varepsilon}\left[\hat{f}(\bs x)\right]^2\right]=&\tau^2 \Psi_3^v,\label{eq:3}\\
    \E_{\bs x,\bs X}\left[\E_{\bs\varepsilon,\bs\Theta}\left[\hat{f}(\bs x)\right]^2\right]=&F^2\Psi_2^e,\label{eq:4}\\
    \E_{\bs x,\bs X} \left[\E_{\bs\varepsilon,\bs\Theta} \left[\hat{f}(\bs x)^2\right]\!-\!\E_{\bs\varepsilon,\bs \Theta}\left[\hat{f}(\bs x)\right]^2\right]=&\tau^2 \Psi_3^e,\label{eq:5}\\
    \E_{\bs x}\left[\mathbb{E}_{\bs\varepsilon,\bs \Theta,\bs X}\left[\hat{f}(\bs x)\right]^2\right]=&F^2\Psi_2^d,\label{eq:6}
\end{align}

\begin{figure}[tb]
    \centering
		\includegraphics[width=0.6\linewidth]{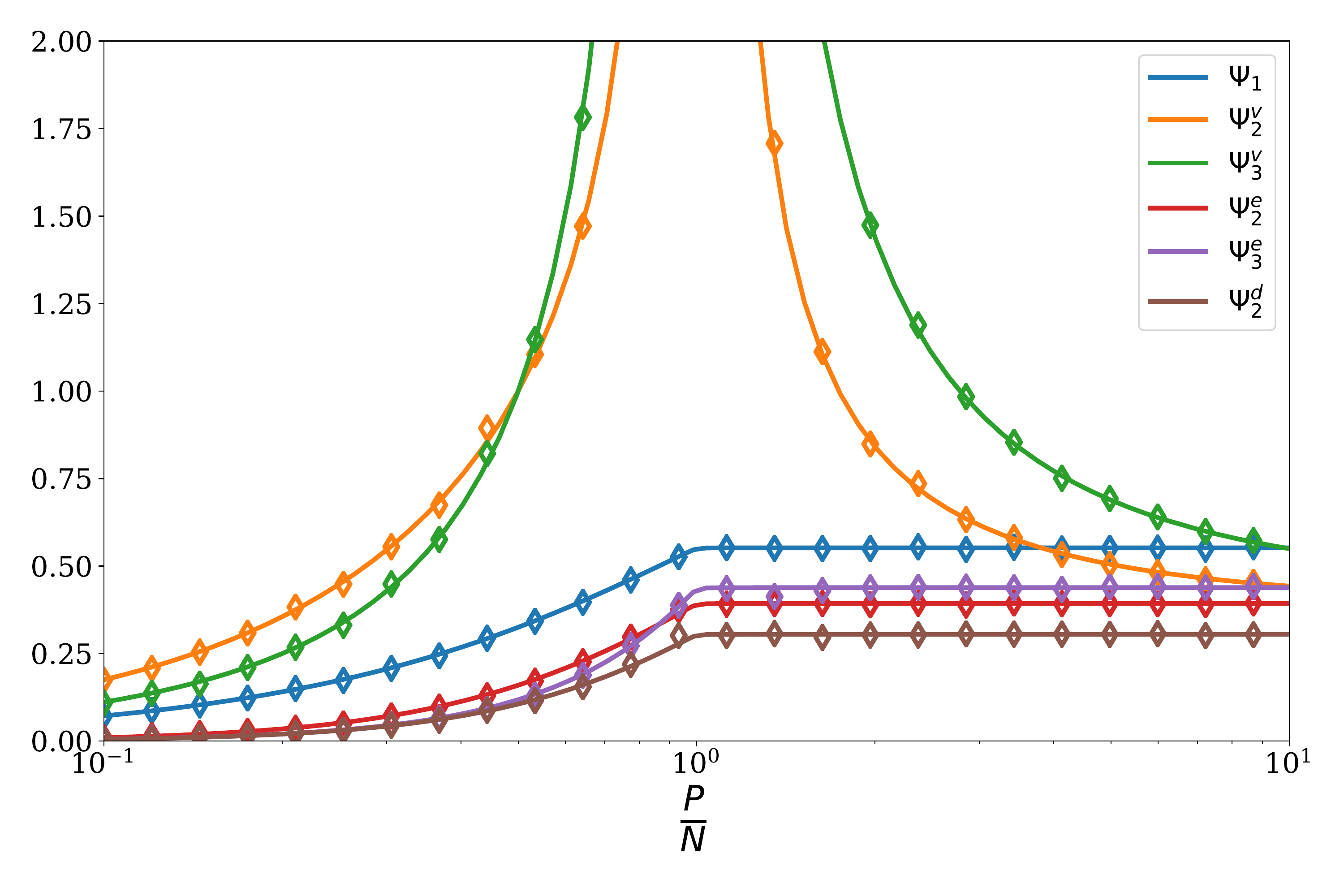}
    \caption{All the terms entering our analytical expressions for the decomposition of the error (Equations \ref{eq:1}-\ref{eq:6}), as function of the overparametrization ratio $P/N$ for $\lambda=10^{-5}$ and $N/D=1$. Numerical estimations obtained for a finite size $D=200$ (diamonds) shows that the asymptotic predictions are extremely accurate even at moderate sizes.}
    \label{fig:termbyterm}
\end{figure}

where the terms $\{\Psi_1,\Psi_2^v,\Psi_3^v,\Psi_2^e,\Psi_3^e,\Psi_2^d\}$, whose full analytical expressions are deferred to Appendix~\ref{app:statement}, are computed following the steps below:

\begin{enumerate}[noitemsep,topsep=0pt,leftmargin=*]
    \item {\it Mapping to a random matrix theory problem.} The first step is to express the right-hand sides of Equations \ref{eq:1}-\ref{eq:6} as traces over random matrices. This is achieved by replacing our model with its asymptotically equivalent Gaussian covariate model \cite{mei2019generalization}, in which the non-linearity of the activation function is encoded as an extra noise term. This enables to take the expectation value with respect to the test sample $\bs x$. 
    \item {\it Mapping to a statistical physics model.} The random matrix theory problem resulting from
    the solution of ridge regression \eqref{eq:estimator} involves inverse random matrices. In order to evaluate their expection value, we use the formula:$$          M^{-1}_{ij} = \lim_{n\to 0} \int \prod_{\alpha=1}^{n}\prod_{i=1}^{D} d\eta_i^\alpha \eta^{1}_i \eta^{1}_j \operatorname{e}^{-\frac{1}{2} \eta_i^\alpha M_{ij} \eta_j^\alpha},$$
    which is based on the Replica Trick~\cite{mezard1987spin,bun2016cleaning}.
    The Gaussian integrals over $\bs \varepsilon, \bs \Theta, \bs X$ can then be straightforwardly performed and lead to a Statistical Physics model for the auxiliary variables $\eta_i^\alpha$.
    \item {\it Mean-Field Theory.} The model for the $\eta_i^\alpha$ variables can then be solved by introducing as order parameters the $n\times n$ overlap matrices $Q^{\alpha\beta} = \frac{1}{P} \sum_{i=1}^P\eta^\alpha_i\eta^\beta_i$ and using replica theory~\cite{mezard1987spin}, see Appendix~\ref{app:computation} for the detailed computation\footnote{In order to obtain the asymptotic formulas for the $\Psi$'s we need to compute (what are called in the Statistical Physics jargon) fluctuations around mean-field theory.}. 
\end{enumerate}

The $\Psi$'s may also be estimated numerically at finite size by evaluating the traces of the random matrices appearing in the Gaussian covariate model at the end of step 1. Figure \ref{fig:termbyterm} shows that results thus obtained are in excellent agreement with the asymptotic expressions even at moderate sizes, e.g. $D=200$, proving the robustness of steps 2 and 3, which differ from the approach presented in~\cite{mei2019generalization}.
 
The indices $v,e,d$ in $\{\Psi_1,\Psi_2^v,\Psi_3^v,\Psi_2^e,\Psi_3^e,\Psi_2^d\}$ stand for \emph{vanilla}, \emph{ensemble} and \emph{divide and conquer}. The \emph{vanilla} terms are sufficient to obtain the test error of a single RF model and were computed in~\cite{mei2019generalization}. The \emph{ensemble} and \emph{divide and conquer} terms allow to obtain the test error obtained when averaging the predictions of several different learners trained respectively on the same dataset and on different splits of the original dataset (see \secref{sec:ensembling}). Figure \ref{fig:termbyterm} shows that the \emph{vanilla} terms exhibit a radically different behavior from the others: at vanishing regularization, they diverge at $P=N$ then decrease monotonically, whereas the others display a kink followed by a plateau. This behavior will be key to the following analysis.

\section{Analysis of Bias and Variances}
\label{sec:decomposition}
\begin{figure}[t!]
    \centering
		\includegraphics[width=0.49\linewidth]{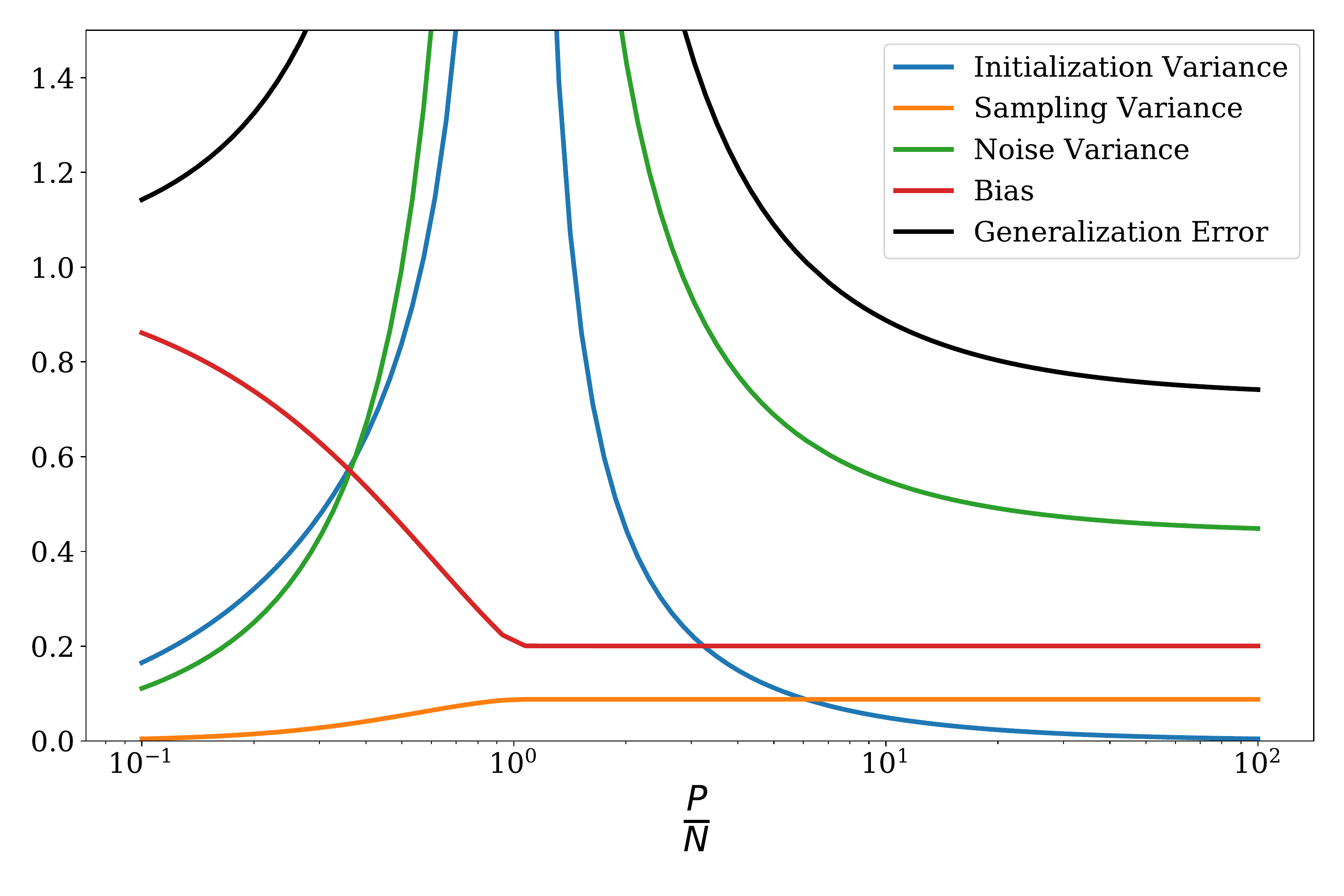}
		\includegraphics[width=0.49\linewidth]{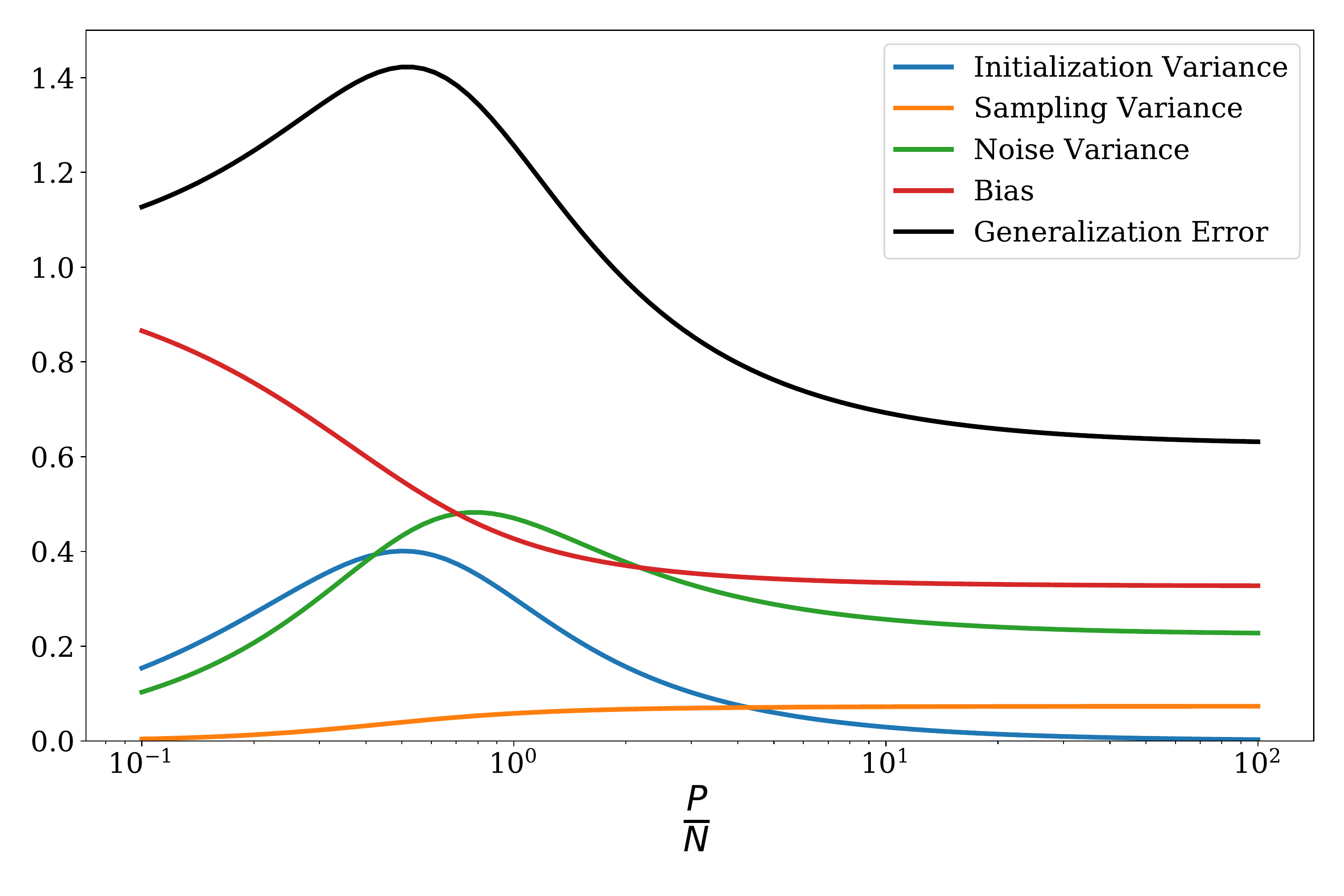}
		\caption{Decomposition of the test error into the bias and the various sources of variance as function of the overparametrization ratio $P/N$ for $N/D=1$, $\mathrm{SNR}=F/\tau=1$. Two values of the regularization constant are used: $\lambda=10^{-5}$ (\textbf{left}) and $\lambda=10^{-1}$ (\textbf{right}). Notice the contrasting behaviors at the interpolation threshold: the noise and initialization variances diverge then decrease monotonically whereas the sampling variance and the bias display a kink followed by a plateau. These singular behaviors are smoothed out by regularization.}
    \label{fig:decomposition}
\end{figure}

The results of the previous section, allow to rewrite the decomposition of the test error as follows:
\begin{align}
    \mathcal{E}_{\rm Noise}&=\tau^2\Psi_3^v,\\
    \mathcal{E}_{\rm Init}&=F^2(\Psi_2^v-\Psi_2^e),\\
    \mathcal{E}_{\rm Samp}&=F^2\left( \Psi_2^e  - \Psi_2^d\right),\\ \mathcal{E}_{\rm Bias}&=F^2 \left(1-2\Psi_1 + \Psi_2^d\right).
\end{align}
These contributions, together with the test error, are shown in \figref{fig:decomposition} in the case of small (top) and large (bottom) regularization.

\paragraph{Interpolation Threshold}
The peak at the interpolation threshold
is completely due to noise and initialization variance, which both diverge at vanishing regularization. In contrast, 
the sampling variance and the bias remain finite and exhibit a phase transition at $P=N$, which is revealed by a kink at vanishing regularization. Adding regularization smooths out these singular behaviours: it removes the divergence and irons out the kink.

\paragraph{Overparametrized regime}
In the overparametrized regime, the sampling variance and the bias do not vary substantially (they remain constant for vanishing regularization). The decrease of the test error is entirely due to the decrease of the noise and initalization variances for $P\!>\!N$. In the limit $P/N\!\to\!\infty$, the initialization variance vanishes, whereas there remains an irreducible noise variance. 

\paragraph{Discussion} In conclusion, we find that the \emph{origin of the double descent curve lies in the behavior of  
noise and initialization variances}. The benefit of overparametrizing stems only from reducing these two contributions.

These results are qualitatively similar to the empirical decomposition of~\cite{neal2018modern} for real neural networks. 
The divergence of the test error as $(P/N-1)^{-1}$ at the interpolation threshold is in agreement with the results of \cite{mei2019generalization}\footnote{Note  that for classification problems the singularity is different~\cite{geiger2019scaling}.}.
As for the decrease of the test error in the over-parametrized regime, we find consistently with the scaling arguments of~\cite{geiger2019scaling} that the initialization error  asymptotically decays to zero inversely proportional to the width (see Appendix~\ref{app:scalings} for more details).
The interpretation of our results differ  from those of  \cite{mei2019generalization} where the authors relate the over-fitting peak occurring at $P=N$ to a divergence in both the variance and the bias terms. This is due to the fact the bias term, as defined in that paper, also includes the initialization variance\footnote{For a given set of random features this is legitimate, but from the perspective of lazy learning the randomness in the features corresponds to the one due to initialization, which is an additional source of variance.}. When the two are disentangled, it becomes clear that it is only the latter which is responsible for the divergence: the bias is, in fact, well-behaved at $P=N$.

\paragraph{Intuition}
The phenomenology described above can be understood by noting that the model
essentially performs linear regression, learning a vector $\bs a \in \mathbb{R}^P$ on a projected dataset $\bs Z\in \mathbb{R}^{N\times P}$ (the activations of the hidden nodes of the RF network). Since $\bs a$ is constrained to lie  in the space spanned by $\bs Z$, which is of dimension $\min(N,P)$, the model gains expressivity when $P$ increases while staying smaller than $N$.

At $P\!=\!N$, the problem becomes fully determined: the data is perfectly interpolated for vanishing $\lambda$. Two types of noise are overfit: (i) the \emph{stochastic noise} corrupting the labels, yielding the divergence in noise variance, and (ii) the \emph{deterministic noise} stemming from the non-linearity of the activation function which cannot be captured, yielding the divergence in initialization variance. However, by further increasing $P$, the noise is spread over more and more random features and is effectively averaged out. Consequently, the test error decreases again as $P$ increases. 

When we make the problem deterministic by averaging out all sources of randomness, i.e. by considering the bias, we see that increasing $P$ beyond $N$ has no effect whatsoever. Indeed, the extra degrees of freedom, which lie in the null space of $\bs Z$, do not provide any extra expressivity: at vanishing regularization, they are killed by the pseudo-inverse to reach the minimum norm solution. For non-vanishing $\lambda$, a similar phenomenology is observed but the interpolation threshold is reached slightly after $P=N$ since the expressivity of the learner is lowered by regularization.

\section{On the effect of ensembling}
\label{sec:ensembling}
In order to further study the effect of the variances on the test error,  we follow \cite{geiger2019scaling} and study the impact of ensembling. In the lazy regime of deep neural networks, the initial values of the weights only affect the gradient at initialization, which corresponds to the vector of random features. Hence, we can study the effect of ensembling in the lazy regime by averaging the predictions of RF models with independently drawn random feature vectors.

\paragraph{Expression of the test error} 

\begin{figure}[htb]
\centering
		\includegraphics[width=0.6\linewidth]{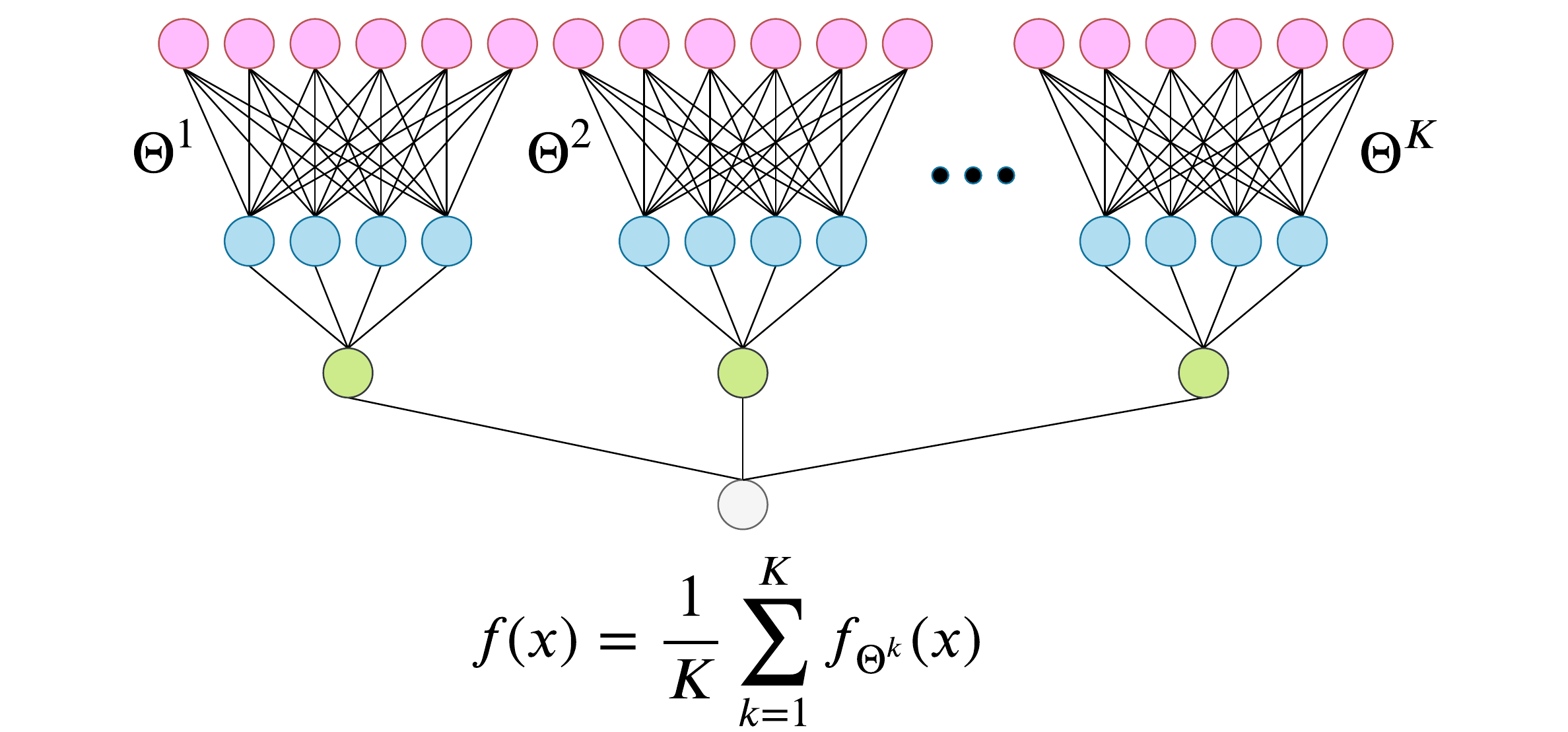}
	\caption{Illustration of the ensembling procedure over $K$ Random Features networks trained on the same data but with different realizations of the first layer, $\{\bs \Theta_1, ... \bs\Theta_K\}$.}
	\label{fig:rf-multiple}
\end{figure}

Consider a set of $K\!>\!1$ RF networks whose first layer weights are drawn independently. These networks are trained independently on the \emph{same} training set. In the analogy outlined above, they correspond to $K$ independent inizializations of the neural network. 
At the end of training, one obtains $K$ estimators $\{\hat{f}_{\bs \Theta^k}\}$ ($k=1,...,K$). When a new sample $\vx$ is presented to the system, the output is taken to be the average over the outputs of the $K$ networks, as illustrated in~\figref{fig:rf-multiple}. By expanding the square and taking the expectation with respect to the random initalizations, the test error can then be written as:
\begin{align}
  \E_{\{\bs \Theta^k\}}\left[\mathcal{R}_{\mathrm{RF}}\right] &= \E_{\substack{\bs x\\ \{\bs \Theta^k\}}} \left[ \left ( \langle \bs \beta, \bs x \rangle - \frac{1}{K} \sum_k \hat f_{\bs\Theta_k}(\bs x) \right)^2 \right]\notag\\
  &= \E_{\bs x} \left[ \langle \bs \beta, \bs x \rangle^2\right]
  - \frac{2}{K} \sum_{i=1}^{K}\E_{\substack{\bs x\\ \bs \Theta_i}} \left[  \langle \bs \beta, \bs x \rangle \hat f_{\bs \Theta_i}(\bs x) \right]+ \frac{1}{K^2} \sum_{i,j=1}^{K} \E_{\substack{\bs x\\ \bs \Theta_i, \bs \Theta_j}} \left[ \hat f_{\bs \Theta_i}(\bs x) \hat f_{\bs \Theta_j}(\bs x) \right].
  \label{eq:ensemble-error-intuition}
\end{align}

The key here is to isolate in the double sum the $K(K-1)$ \emph{ensemble} terms $i\neq j$, which involve two different initalizations and yield $\E_{\bs x} \left[ \E_{\bs \Theta}\left[\hat f_{\bs \Theta}(\bs x)\right]^2 \right]$, from the $K$ \emph{vanilla} terms which give $\E_{\bs x, \bs\Theta} \left[ \hat f_{\bs \Theta}(\bs x)^2 \right]$. This allows to express the test error in terms of the quantities defined in \eqref{eq:1} to \eqref{eq:6} and leads to the analytic formula for the test error valid for any $K\in \mathbb{N}$:
\begin{align}
   \E_{\{\bs \Theta^{(k)}\}, \bs X, \boldsymbol{\varepsilon}} \left[ \mathcal{R}_{\mathrm{RF}}\right]&=F^2 \left(1-2\Psi_1^v\right)+\frac{1}{K}\left(F^2\Psi_2^v+ \tau^2\Psi_3^v \right)+\left(1\!-\!\frac{1}{K}\right)\! \left(F^2\Psi_2^e \!+\! \tau^2\Psi_3^e \right).
   \label{eq:ensemble-error}
\end{align}
We see that ensembling amounts to a linear interpolation between the \emph{vanilla} terms $\Psi_2^v, \Psi_3^v$, for $K=1$, and the \emph{ensemble} terms $\Psi_2^e, \Psi_3^e$ for $K\to \infty$. 

\begin{figure}[htb]
  \centering
  		\includegraphics[width=0.6\linewidth]{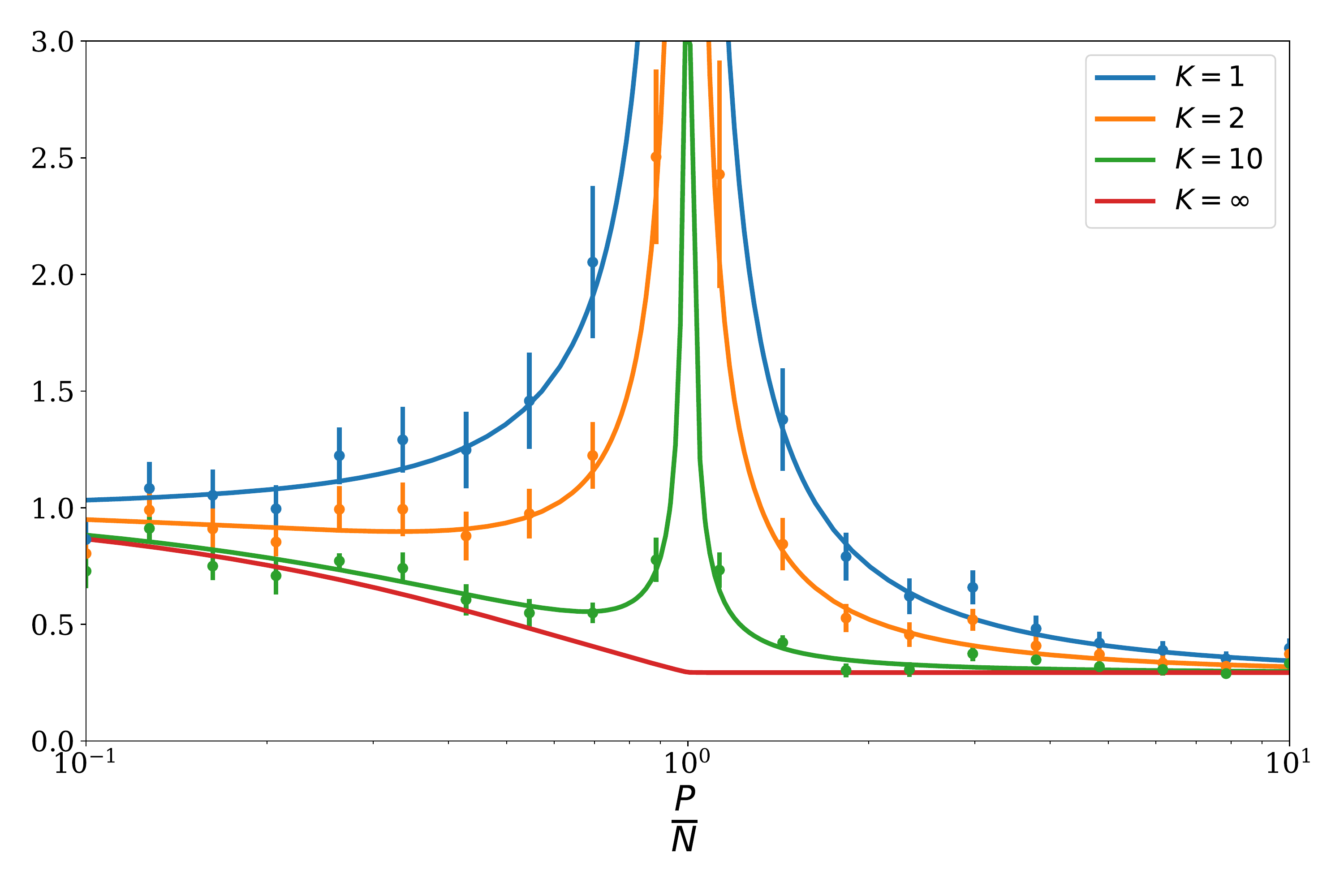}
    \vspace{-0.5cm}
	\caption{Test error when ensembling $K=1,2,10$ differently initialized RF models as function of the overparametrization ratio $P/N$. We fixed $\lambda=10^{-5}$, $N/D=1$, ${\mathrm SNR}=10$. For comparison, we show the results of numerical simulations at finite $D=200$: the vertical bars depict the standard deviation over $10$ runs. The variability observed here was absent in~\figref{fig:termbyterm} because we are considering the true RF model rather than the asymptotically equivalent Gaussian covariate model. This shows that most of this variability is caused by the finite-size deviation between the two models.	Note that our analytic expression~\eqref{eq:ensemble-error} gives us access to the limit $N\to\infty$, where the divergence at $P=N$ is entirely suppressed.}
  \label{fig:double-descent}
\end{figure}

The effect of ensembling on the double descent curve is shown in \figref{fig:double-descent}. As $K$ increases, the overfitting peak at the interpolation threshold is diminished. This observation is very similar to the empirical findings of~\cite{geiger2019scaling} in the context of real neural networks. Our analytic expression agrees with the numerical results obtained by training RF models, even at moderate size $D=200$.

Note that a related procedure is the \emph{divide and conquer} approach, where the dataset is partitioned into $K$ splits of equal size and each one of the $K$ differently initialized learners is trained on a distinct split. This approach was studied for kernel learning in~\cite{drucker1994boosting}, and is analyzed within our framework in Appendix~\ref{app:divide-and-conquer}.

\paragraph{Ensembling reduces the double trouble} 

The bias-variance decomposition of the test
error makes the suppression of the divergence explicit. The bias and variances contribution read for the averaged estimator:
\begin{align}
  \mathcal{E}_{\rm Noise}&=\tau^2\left(\Psi_3^e+\frac{1}{K}\left(\Psi_3^v-\Psi_3^e\right)\right),\\
  \mathcal{E}_{\rm Init}&=\frac{F^2}{K}\left(\Psi_2^v-\Psi_2^e\right),\\
  \mathcal{E}_{\rm Samp}&=F^2\left( \Psi_2^e - \Psi_2^d\right),\\ \mathcal{E}_{\rm Bias}&=F^2 \left(1-2\Psi_1 + \Psi_2^d\right).
\end{align}

\begin{figure}[tb]
\centering
		\includegraphics[width=0.6\linewidth]{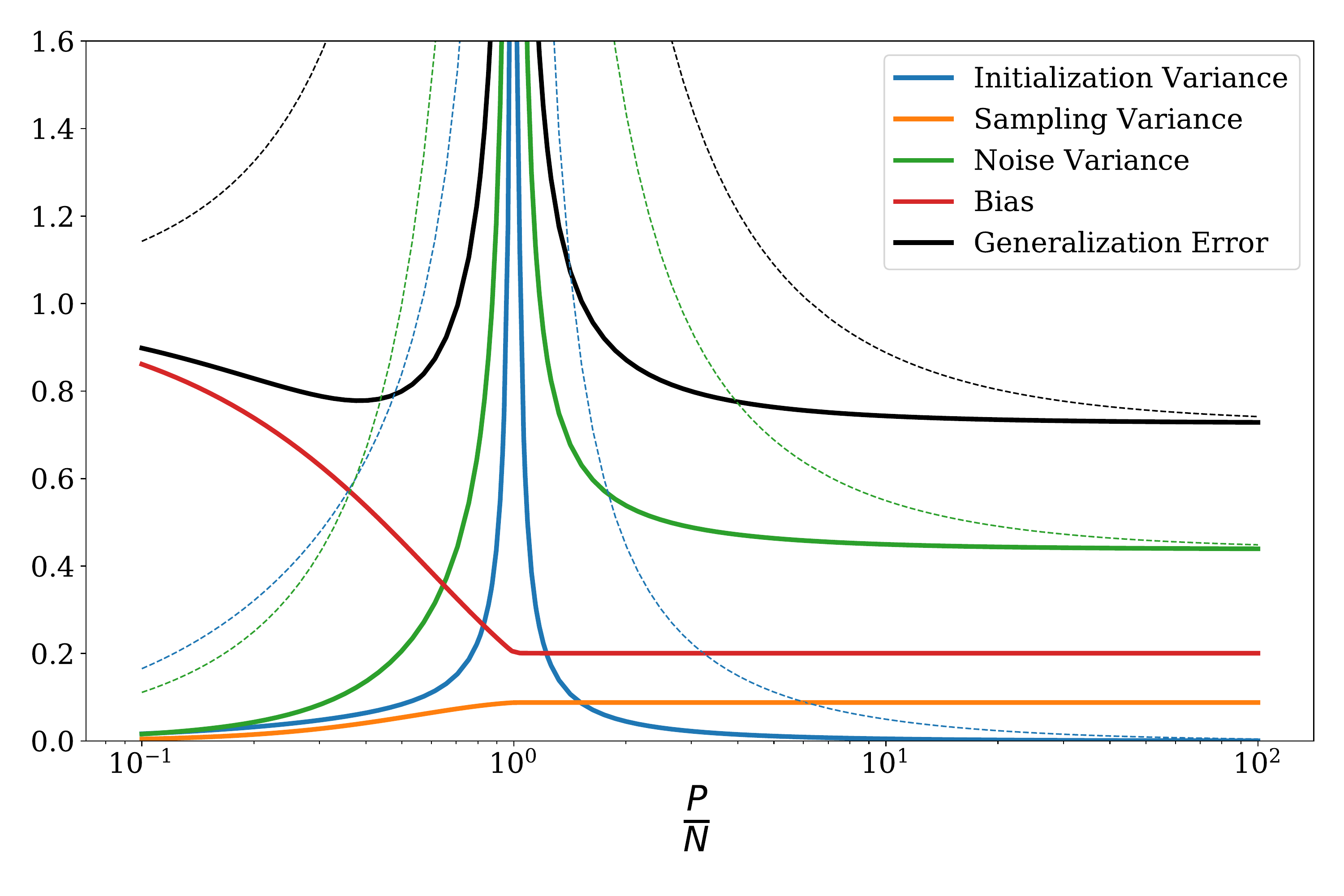}
    \vspace{-0.5cm}
	\caption{
	Decomposition of the test error into the bias and the various sources of variance as function of the overparametrization ratio $P/N$ for $\lambda=10^{-5}$, $N/D=1$, $\mathrm{SNR}=1$. The thin dashed lines are taken from \figref{fig:decomposition} (top) where we had $K=1$; the thick solid lines show how ensembling at $K=10$ suppressed the divergences of the noise and initialization variances.}
	\label{fig:multiple_k_decomposition}
\end{figure}

These equations show that ensembling only affects the noise and initialization variances. In both cases, their divergence at the interpolation threshold (due to $\Psi_2^v,\Psi_3^v$) is suppressed as $1/K$, see \figref{fig:multiple_k_decomposition} for an illustration.
At $P\!>\!N$, ensembling and overparametrizing have a very similar effect: they wipe out these two troubling sources of randomness by averaging them out over more random features. Indeed, we see in \figref{fig:double-descent} that in this overparametrized regime, sending $K\!\to\!\infty$ has the same effect as sending $P/N\!\to\!\infty$: in both cases the system approaches the kernel limit. At $P\!<\!N$, this is not true: as shown in~\cite{jacot2020implicit}, the $K\!\to\!\infty$ predictor still operates in the kernel limit, but with an effective regularization parameter $\tilde\lambda>\lambda$ which diverges as $P/N\!\to\!0$. This (detrimental) implicit regularization increases the generalization error.

\begin{figure}[h!]
  \centering
	\includegraphics[width=0.6\linewidth]{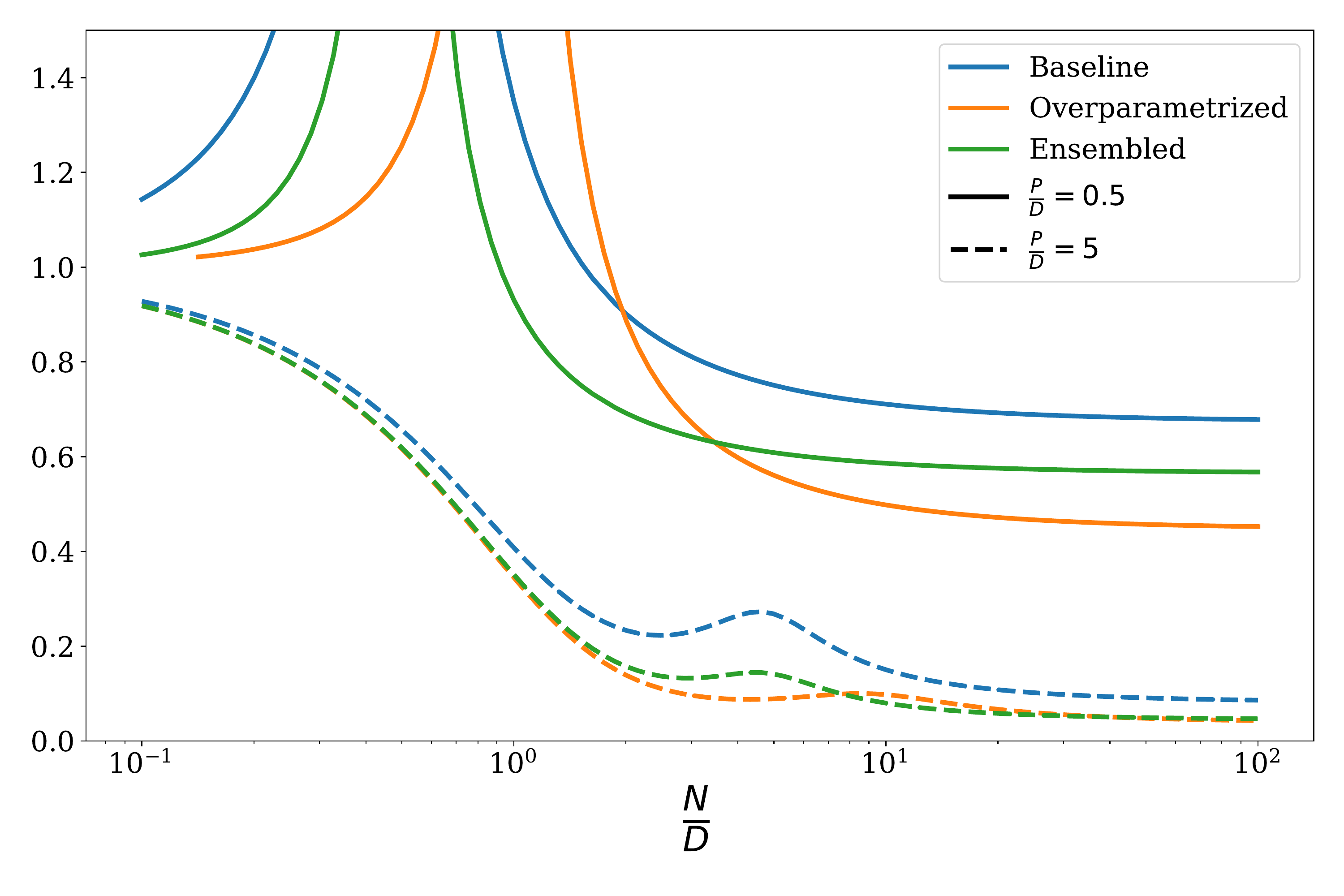}
    \vspace{-0.5cm}
	\caption{Comparison of the test error of a RF model (blue) with that obtained by doubling the number features (orange) or ensembling over two initializations of the features (green), as function of $N/D$. The parameters are $\lambda=10^{-5}$, $\mathrm{SNR}=10$, $P/D=0.5$ (solid lines) and $P/D=5$ (dashed lines).}
  \label{fig:ensemble-vs-overparametrize}
\end{figure}

\paragraph{Ensembling vs. overparametrization}
As we have shown, ensembling and overparametrizing have similar effects in the lazy regime. But which is more powerful: ensembling $K$ models, or using a single model with $K$ times more features? The answer 
is given in \figref{fig:ensemble-vs-overparametrize} for $K=2$ where 
we plot our analytical results while varying the number of data points, $N$. Two observations are particularly interesting. First, overparametrization shifts the interpolation threshold, opening up a region where ensembling outperforms overparametrizing.
Second, overparametrization yields a higher asymptotic improvement in the large dataset limit $N/D\to \infty$, but the gap between overparametrizing and ensembling is reduced as $P/D$ increases. At $P\!\gg\! D$, where we are already close to the kernel limit, both methods yield a similar improvement. Note that from the point of view of efficiency, ridge regression involves the inversion of a $P\times P$ matrix, therefore ensembling is significantly more efficient.

\paragraph{Ensembling vs. optimal regularization}

\begin{figure}[h!]
		\includegraphics[width=\linewidth]{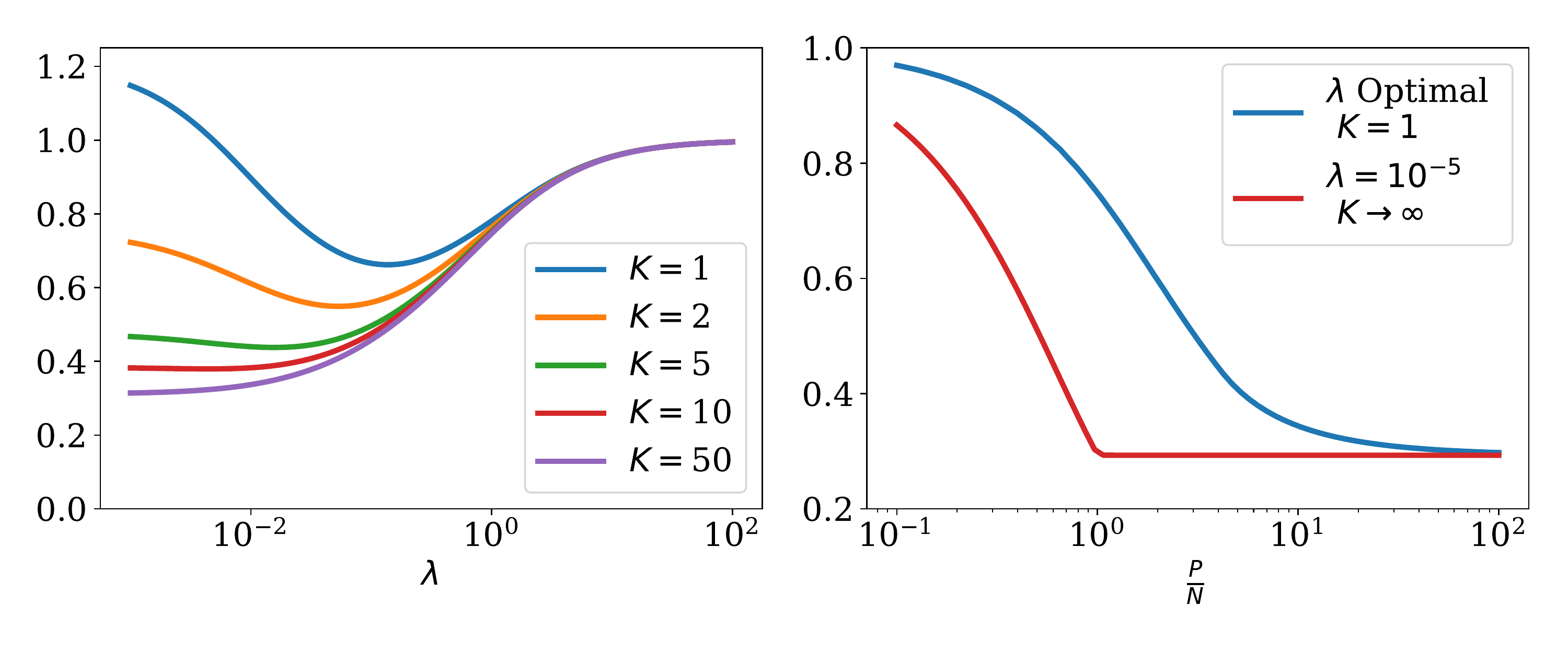}  
		\vspace{-0.8cm}
			\caption{\textbf{Left:} Test error as a function of $\lambda$ for various values of $K$ and parameters $P/D=2$, $N/D=1$, $\mathrm{SNR}=10$. \textbf{Right:} Comparison of test error for an optimal regularized system with $K=1$ and the system with $K\to\infty$ with $\lambda=10^{-5}$. Optimization performed over 50 values of $\lambda$ from $10^{-5}$ to $10^2$. Parameters are $N/D=1$, $\mathrm{SNR}=10$.}
  \label{fig:ensemble-vs-regularize}
\end{figure}

In all the results presented above, we keep the regularization constant $\lambda$ fixed. However, by appropriately choosing the value of $\lambda$ at each value of $P/N$, the performance is improved. As \figref{fig:ensemble-vs-overparametrize} (left) reveals, the optimal value of $\lambda$ decreases with $K$ since the minimum of the test error shifts to the left when increasing $K$. In other words, ensembling is best when the predictors one ensembles upon are individually under-regularized, as was observed previously for kernel learning in~\cite{zhang2015divide}. Figure~\ref{fig:ensemble-vs-regularize} (right) shows that an infinitely ensembled model ($K\to\infty$) always performs better than an optimally regularized single model ($K=1$).

\section{Numerical experiments on neural networks}
\label{sec:numerics}
\begin{figure}[h!]
    \centering
		\includegraphics[width=\linewidth]{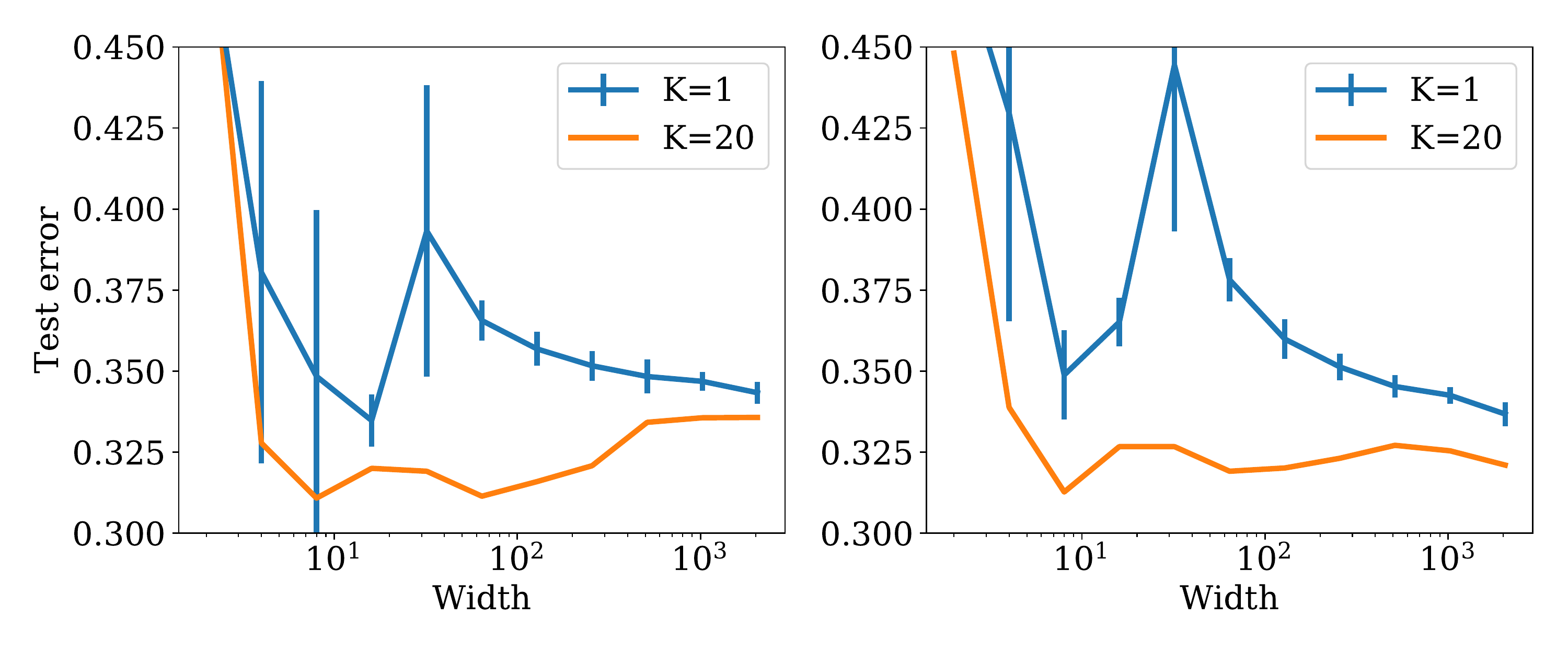}
\vspace{-0.8cm}
	\caption{Test error on the binary 10-PCA CIFAR10 as function of the number of nodes per layer of the 5-layer neural network trained until convergence with the full-batch \emph{Adam}. We compare the test error of a single predictor ($K=1$), averaged over $20$ initializations of the weights (the standard deviation is depicted as vertical bars), with the ensembling predictor at $K=20$. \textbf{Left:} $\alpha=10$. \textbf{Right:} $\alpha=100$, where we are closer to the lazy regime and the ensembling curve flattens beyond the interpolation threshold, which occurs around 30 nodes per layer.}
    \label{fig:lazy}
\end{figure}

Finally, we investigate whether the phenomenology described here holds for realistic neural networks learning real data in the lazy regime. We follow here the protocol used in~\cite{geiger2019scaling,geiger2019disentangling} and  train a 5-layer fully-connected network on the CIFAR-10 dataset. We keep only the first ten PCA components of the images, and divide the images in two classes according to the parity of the labels. We perform $10^5$ steps of full-batch gradient descent with the \emph{Adam} optimizer and a learning rate of $0.1$, and scale the weights as prescribed in~\cite{jacot2018neural}.

We gradually go from the usual \emph{feature learning} regime to the \emph{lazy learning} regime using the trick introduced in~\cite{chizat2018note}, which consists in scaling the output of the network by a factor $\alpha$ and replacing the learning function $f_\theta(\bs x)$ by $\alpha (f_\theta(\bs x)\!-\!f_{\theta_0}(\bs x))$. For $\alpha\!\gg\! 1$, one must have that $\theta\!-\!\theta_0\! \sim\! {1}/{\alpha}$ in order for the learning function to remain of order one. In other words, the weights are forced to stay close to their initialization, hence the name \emph{lazy learning}. 

Results are shown in \figref{fig:lazy}. Close to the lazy regime ($\alpha\!=\!100$, right panel), a very similar behavior as the RF model is observed. The test error curve\footnote{Note that we are considering a binary classification task here: the error is defined as the fraction of misclassified images.} obtained when ensembling $K=20$ independently initialized networks becomes roughly flat after the interpolation threshold (which here is signalled by the peak in the test accuracy). As we move away from the lazy regime ($\alpha\!=\!10$, left panel), the same curve develops a \emph{dip} around the interpolation threshold and increases beyond $P\!>\!N$ as observed previously in~\cite{geiger2019scaling}. This may arguably be associated to the beneficial effect of \emph{feature learning}, as discussed in \cite{geiger2019disentangling} where the transition from lazy to feature learning was investigated.

\paragraph{Acknowledgements}
We thank Matthieu Wyart and Lenka Zdeborov\'a for discussions related to this project. 
This work is supported by the French Agence Nationale de la Recherche under grant
ANR-17-CE23-0023-01 PAIL and ANR-19-P3IA-0001 PRAIRIE, and by the Simons Foundation ($\#$454935, Giulio Biroli). We also acknowledge support from the chaire CFM-ENS  ``Science des donn\'ees''.

Upon completion of this paper, we became aware of two related parallel works presenting bias-variance tradeoffs for RF models. The variance due to the sampling of the dataset was considered in~\cite{yang2020rethinking}, whereas \cite{jacot2020implicit} focused on the variance due to the randomness of the random feature vectors. 

\newpage
\bibliographystyle{unsrt}
\bibliography{refs}

\begin{thebibliography}{10}

\bibitem{geiger2019scaling}
Mario Geiger, Arthur Jacot, Stefano Spigler, Franck Gabriel, Levent Sagun,
  St{\'e}phane d'Ascoli, Giulio Biroli, Cl{\'e}ment Hongler, and Matthieu
  Wyart.
\newblock Scaling description of generalization with number of parameters in
  deep learning.
\newblock {\em arXiv preprint arXiv:1901.01608}, 2019.

\bibitem{krizhevsky2012imagenet}
Alex Krizhevsky, Ilya Sutskever, and Geoffrey~E Hinton.
\newblock Imagenet classification with deep convolutional neural networks.
\newblock In {\em Advances in neural information processing systems}, pages
  1097--1105, 2012.

\bibitem{lecun2015deep}
Yann LeCun, Yoshua Bengio, and Geoffrey Hinton.
\newblock Deep learning.
\newblock {\em nature}, 521(7553):436--444, 2015.

\bibitem{hinton2012deep}
Geoffrey Hinton, Li~Deng, Dong Yu, George~E Dahl, Abdel-rahman Mohamed, Navdeep
  Jaitly, Andrew Senior, Vincent Vanhoucke, Patrick Nguyen, Tara~N Sainath,
  et~al.
\newblock Deep neural networks for acoustic modeling in speech recognition: The
  shared views of four research groups.
\newblock {\em IEEE Signal processing magazine}, 29(6):82--97, 2012.

\bibitem{sutskever2014sequence}
Ilya Sutskever, Oriol Vinyals, and Quoc~V Le.
\newblock Sequence to sequence learning with neural networks.
\newblock In {\em Advances in neural information processing systems}, pages
  3104--3112, 2014.

\bibitem{zhang2016understanding}
Chiyuan Zhang, Samy Bengio, Moritz Hardt, Benjamin Recht, and Oriol Vinyals.
\newblock Understanding deep learning requires rethinking generalization.
\newblock {\em arXiv preprint arXiv:1611.03530}, 2016.

\bibitem{advani2017high}
Madhu~S Advani and Andrew~M Saxe.
\newblock High-dimensional dynamics of generalization error in neural networks.
\newblock {\em arXiv preprint arXiv:1710.03667}, 2017.

\bibitem{belkin2018reconciling}
Mikhail Belkin, Daniel Hsu, Siyuan Ma, and Soumik Mandal.
\newblock Reconciling modern machine learning and the bias-variance trade-off.
\newblock {\em arXiv preprint arXiv:1812.11118}, 2018.

\bibitem{neal2018modern}
Brady Neal, Sarthak Mittal, Aristide Baratin, Vinayak Tantia, Matthew Scicluna,
  Simon Lacoste-Julien, and Ioannis Mitliagkas.
\newblock A modern take on the bias-variance tradeoff in neural networks.
\newblock {\em arXiv preprint arXiv:1810.08591}, 2018.

\bibitem{hastie2019surprises}
Trevor Hastie, Andrea Montanari, Saharon Rosset, and Ryan~J Tibshirani.
\newblock Surprises in high-dimensional ridgeless least squares interpolation.
\newblock {\em arXiv preprint arXiv:1903.08560}, 2019.

\bibitem{mei2019generalization}
Song Mei and Andrea Montanari.
\newblock The generalization error of random features regression: Precise
  asymptotics and double descent curve.
\newblock {\em arXiv preprint arXiv:1908.05355}, 2019.

\bibitem{neyshabur2014search}
Behnam Neyshabur, Ryota Tomioka, and Nathan Srebro.
\newblock In search of the real inductive bias: On the role of implicit
  regularization in deep learning.
\newblock {\em arXiv preprint arXiv:1412.6614}, 2014.

\bibitem{spigler2018jamming}
Stefano Spigler, Mario Geiger, St{\'e}phane d'Ascoli, Levent Sagun, Giulio
  Biroli, and Matthieu Wyart.
\newblock A jamming transition from under-to over-parametrization affects loss
  landscape and generalization.
\newblock {\em arXiv preprint arXiv:1810.09665}, 2018.

\bibitem{nakkiran2019deep}
Preetum Nakkiran, Gal Kaplun, Yamini Bansal, Tristan Yang, Boaz Barak, and Ilya
  Sutskever.
\newblock Deep double descent: Where bigger models and more data hurt.
\newblock {\em arXiv preprint arXiv:1912.02292}, 2019.

\bibitem{breiman1995reflections}
Leo Breiman.
\newblock Reflections after refereeing papers for nips.
\newblock {\em The Mathematics of Generalization}, pages 11--15, 1995.

\bibitem{opper1996statistical}
Manfred Opper and Wolfgang Kinzel.
\newblock Statistical mechanics of generalization.
\newblock In {\em Models of neural networks III}, pages 151--209. Springer,
  1996.

\bibitem{liu1998jamming}
Andrea~J Liu and Sidney~R Nagel.
\newblock Jamming is not just cool any more.
\newblock {\em Nature}, 396(6706):21--22, 1998.

\bibitem{engel2001statistical}
Andreas Engel and Christian Van~den Broeck.
\newblock {\em Statistical mechanics of learning}.
\newblock Cambridge University Press, 2001.

\bibitem{franz2016simplest}
Silvio Franz and Giorgio Parisi.
\newblock The simplest model of jamming.
\newblock {\em Journal of Physics A: Mathematical and Theoretical},
  49(14):145001, 2016.

\bibitem{krzakala2007landscape}
Florent Krzakala and Jorge Kurchan.
\newblock Landscape analysis of constraint satisfaction problems.
\newblock {\em Physical Review E}, 76(2):021122, 2007.

\bibitem{zdeborova2007phase}
Lenka Zdeborov{\'a} and Florent Krzakala.
\newblock Phase transitions in the coloring of random graphs.
\newblock {\em Physical Review E}, 76(3):031131, 2007.

\bibitem{rosset2004margin}
Saharon Rosset, Ji~Zhu, and Trevor~J Hastie.
\newblock Margin maximizing loss functions.
\newblock In {\em Advances in neural information processing systems}, pages
  1237--1244, 2004.

\bibitem{chizat2018note}
L\'{e}na\"{\i}c Chizat, Edouard Oyallon, and Francis Bach.
\newblock On lazy training in differentiable programming.
\newblock In H.~Wallach, H.~Larochelle, A.~Beygelzimer, F.~d'Alch\'{e} Buc,
  E.~Fox, and R.~Garnett, editors, {\em Advances in Neural Information
  Processing Systems 32}, pages 2933--2943. Curran Associates, Inc., 2019.

\bibitem{arora2019exact}
Sanjeev Arora, Simon~S Du, Wei Hu, Zhiyuan Li, Ruslan Salakhutdinov, and
  Ruosong Wang.
\newblock On exact computation with an infinitely wide neural net.
\newblock {\em arXiv preprint arXiv:1904.11955}, 2019.

\bibitem{mei2019mean}
Song Mei, Theodor Misiakiewicz, and Andrea Montanari.
\newblock Mean-field theory of two-layers neural networks: dimension-free
  bounds and kernel limit.
\newblock {\em arXiv preprint arXiv:1902.06015}, 2019.

\bibitem{woodworth2019kernel}
Blake Woodworth, Suriya Gunasekar, Jason Lee, Daniel Soudry, and Nathan Srebro.
\newblock Kernel and deep regimes in overparametrized models.
\newblock {\em arXiv preprint arXiv:1906.05827}, 2019.

\bibitem{geiger2019disentangling}
Mario Geiger, Stefano Spigler, Arthur Jacot, and Matthieu Wyart.
\newblock Disentangling feature and lazy learning in deep neural networks: an
  empirical study.
\newblock {\em arXiv preprint arXiv:1906.08034}, 2019.

\bibitem{arora2019harnessing}
Sanjeev Arora, Simon~S Du, Zhiyuan Li, Ruslan Salakhutdinov, Ruosong Wang, and
  Dingli Yu.
\newblock Harnessing the power of infinitely wide deep nets on small-data
  tasks.
\newblock {\em arXiv preprint arXiv:1910.01663}, 2019.

\bibitem{daniely2016toward}
Amit Daniely, Roy Frostig, and Yoram Singer.
\newblock Toward deeper understanding of neural networks: The power of
  initialization and a dual view on expressivity.
\newblock In {\em Advances In Neural Information Processing Systems}, pages
  2253--2261, 2016.

\bibitem{lee2017deep}
Jaehoon Lee, Yasaman Bahri, Roman Novak, Samuel~S Schoenholz, Jeffrey
  Pennington, and Jascha Sohl-Dickstein.
\newblock Deep neural networks as gaussian processes.
\newblock {\em arXiv preprint arXiv:1711.00165}, 2017.

\bibitem{jacot2018neural}
Arthur Jacot, Franck Gabriel, and Cl{\'e}ment Hongler.
\newblock Neural tangent kernel: Convergence and generalization in neural
  networks.
\newblock In {\em Advances in neural information processing systems}, pages
  8571--8580, 2018.

\bibitem{rahimi2008random}
Ali Rahimi and Benjamin Recht.
\newblock Random features for large-scale kernel machines.
\newblock In {\em Advances in neural information processing systems}, pages
  1177--1184, 2008.

\bibitem{neyshabur2017geometry}
Behnam Neyshabur, Ryota Tomioka, Ruslan Salakhutdinov, and Nathan Srebro.
\newblock Geometry of optimization and implicit regularization in deep
  learning.
\newblock {\em arXiv preprint arXiv:1705.03071}, 2017.

\bibitem{mezard1987spin}
Marc M{\'e}zard, Giorgio Parisi, and Miguel Virasoro.
\newblock {\em Spin glass theory and beyond: An Introduction to the Replica
  Method and Its Applications}, volume~9.
\newblock World Scientific Publishing Company, 1987.

\bibitem{seung1992statistical}
Hyunjune~Sebastian Seung, Haim Sompolinsky, and Naftali Tishby.
\newblock Statistical mechanics of learning from examples.
\newblock {\em Physical review A}, 45(8):6056, 1992.

\bibitem{advani2013statistical}
Madhu Advani, Subhaneil Lahiri, and Surya Ganguli.
\newblock Statistical mechanics of complex neural systems and high dimensional
  data.
\newblock {\em Journal of Statistical Mechanics: Theory and Experiment},
  2013(03):P03014, 2013.

\bibitem{zdeborova2016statistical}
Lenka Zdeborov{\'a} and Florent Krzakala.
\newblock Statistical physics of inference: Thresholds and algorithms.
\newblock {\em Advances in Physics}, 65(5):453--552, 2016.

\bibitem{livan2018introduction}
Giacomo Livan, Marcel Novaes, and Pierpaolo Vivo.
\newblock {\em Introduction to random matrices: theory and practice},
  volume~26.
\newblock Springer, 2018.

\bibitem{tarquini2016level}
Elena Tarquini, Giulio Biroli, and Marco Tarzia.
\newblock Level statistics and localization transitions of levy matrices.
\newblock {\em Physical review letters}, 116(1):010601, 2016.

\bibitem{aggarwal2018goe}
Amol Aggarwal, Patrick Lopatto, and Horng-Tzer Yau.
\newblock Goe statistics for levy matrices.
\newblock {\em arXiv preprint arXiv:1806.07363}, 2018.

\bibitem{nakkiran2019more}
Preetum Nakkiran.
\newblock More data can hurt for linear regression: Sample-wise double descent.
\newblock {\em arXiv preprint arXiv:1912.07242}, 2019.

\bibitem{belkin2019two}
Mikhail Belkin, Daniel Hsu, and Ji~Xu.
\newblock Two models of double descent for weak features.
\newblock {\em arXiv preprint arXiv:1903.07571}, 2019.

\bibitem{deng2019model}
Zeyu Deng, Abla Kammoun, and Christos Thrampoulidis.
\newblock A model of double descent for high-dimensional binary linear
  classification.
\newblock {\em arXiv preprint arXiv:1911.05822}, 2019.

\bibitem{kini2020analytic}
Ganesh Kini and Christos Thrampoulidis.
\newblock Analytic study of double descent in binary classification: The impact
  of loss.
\newblock {\em arXiv preprint arXiv:2001.11572}, 2020.

\bibitem{replica2020}
Florent Krzakala Marc M\'ezard Lenka~Zdeborov\'a Federica~Gerace,
  Bruno~Loureiro.
\newblock Generalisation error in learning with random features and the hidden
  manifold model.
\newblock {\em arXiv:2002.09339}, 2020.

\bibitem{drucker1994boosting}
Harris Drucker, Corinna Cortes, Lawrence~D Jackel, Yann LeCun, and Vladimir
  Vapnik.
\newblock Boosting and other ensemble methods.
\newblock {\em Neural Computation}, 6(6):1289--1301, 1994.

\bibitem{zhang2013divide}
Yuchen Zhang, John Duchi, and Martin Wainwright.
\newblock Divide and conquer kernel ridge regression.
\newblock In {\em Conference on Learning Theory}, pages 592--617, 2013.

\bibitem{goldt2019modelling}
Sebastian Goldt, Marc M{\'e}zard, Florent Krzakala, and Lenka Zdeborov{\'a}.
\newblock Modelling the influence of data structure on learning in neural
  networks.
\newblock {\em arXiv preprint arXiv:1909.11500}, 2019.

\bibitem{bun2016cleaning}
Jo{\"e}l Bun, Jean-Philippe Bouchaud, and Marc Potters.
\newblock Cleaning correlation matrices.
\newblock {\em Risk magazine}, 2015, 2016.

\bibitem{jacot2020implicit}
Arthur Jacot, Berfin {\c{S}}im{\c{s}}ek, Francesco Spadaro, Cl{\'e}ment
  Hongler, and Franck Gabriel.
\newblock Implicit regularization of random feature models.
\newblock {\em arXiv preprint arXiv:2002.08404}, 2020.

\bibitem{zhang2015divide}
Yuchen Zhang, John Duchi, and Martin Wainwright.
\newblock Divide and conquer kernel ridge regression: A distributed algorithm
  with minimax optimal rates.
\newblock {\em The Journal of Machine Learning Research}, 16(1):3299--3340,
  2015.

\bibitem{yang2020rethinking}
Zitong Yang, Yaodong Yu, Chong You, Jacob Steinhardt, and Yi~Ma.
\newblock Rethinking bias-variance trade-off for generalization of neural
  networks.
\newblock {\em arXiv preprint arXiv:2002.11328}, 2020.

\bibitem{chawla2003distributed}
Nitesh~V Chawla, Thomas~E Moore, Lawrence~O Hall, Kevin~W Bowyer, W~Philip
  Kegelmeyer, and Clayton Springer.
\newblock Distributed learning with bagging-like performance.
\newblock {\em Pattern recognition letters}, 24(1-3):455--471, 2003.

\end{thebibliography}

\newpage
\appendix
\setlength{\parindent}{0pt}

\section{Further analytical results}

\subsection{Asymptotic scalings}
\label{app:scalings}
\begin{figure}[htb]
\centering
		\includegraphics[width=0.9\linewidth]{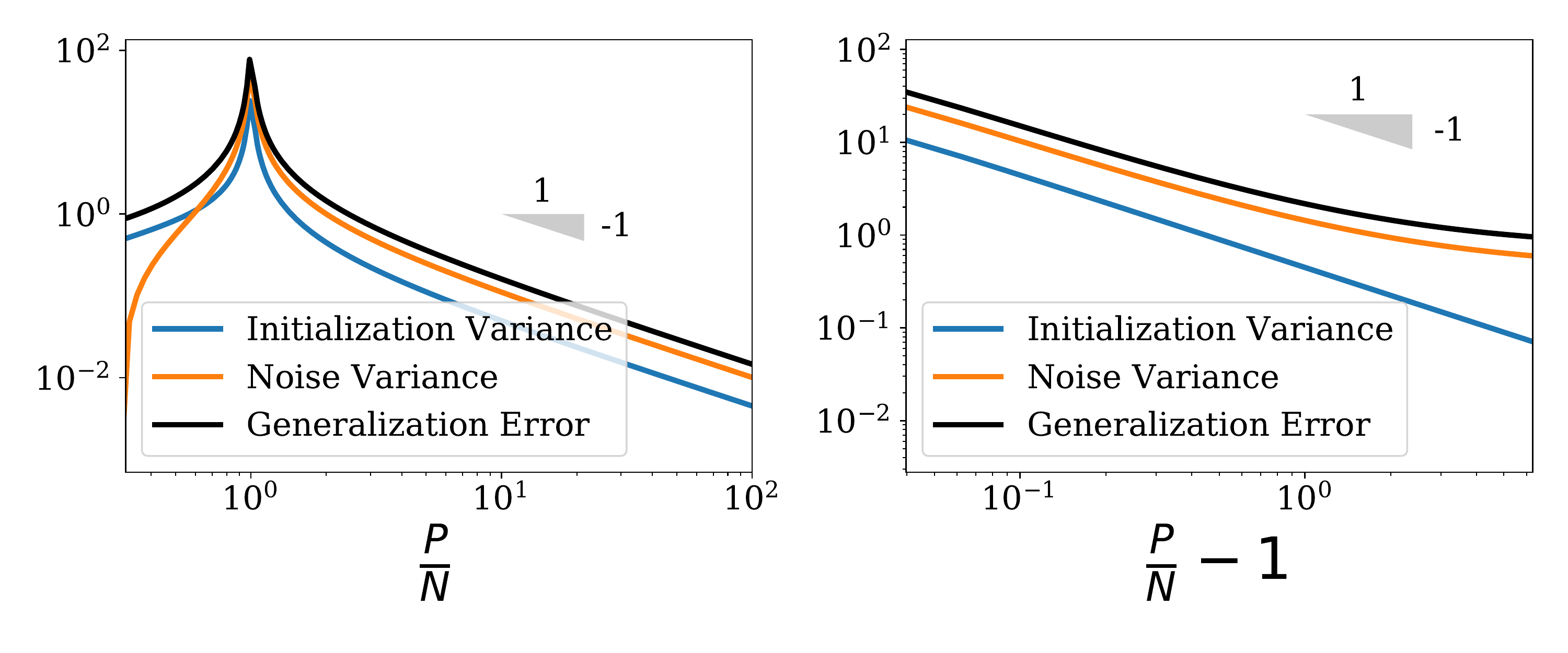}
		\vspace{-0.5cm}
			\caption{Log-log behaviour of the quantities of interest at \textbf{Left:} $P/N\to\infty$ and \textbf{Right}: $P\to N$ with $\lambda=10^{-5}$, $N/D=1$ and $\tau=1$. In both cases, one observes an inverse scaling law.}
   \label{fig:asymptotics}
\end{figure}
Figure~\ref{fig:asymptotics} (left) shows that the various terms entering the decomposition of the generalization error approach their asymptotic values at a rate $(P/N)^{-1}$. This scaling law is consistent with that found in~\cite{geiger2019scaling} for real neural networks, where $P$ is replaced by the width of the layers of the network. As for the divergence of the noise and initialization variances observed at the interpolation threshold, figure~\ref{fig:asymptotics} (right) shows that they also follow an inverse power law $(P/N-1)^{-1}$ at vanishing regularization. 

\subsection{Divide and Conquer approach}
\label{app:divide-and-conquer}

As mentioned in the main text, another way to average the predictions of differently initialized learners is the \emph{divide and conquer} approach ~\cite{drucker1994boosting}. In this framework, the data set is divided into $K$ splits of size $N/K$. Each of the $K$ differently initalized learner is trained on a distinct split. This approach is extremely useful for kernel learning~\cite{zhang2015divide}, where the computational burden is in the inversion of the Gram matrix which is of size $N\times N$. In the random projection approach considered here, it does not offer any computational gain, however it is interesting how it affects the generalization error.

Within our framework, the generalization error can easily be calculated as:
\begin{align}
  \E_{\{\bs \Theta^{(k)}\}, \bs X, \boldsymbol{\varepsilon}} \left[ \mathcal{R}_{\mathrm{RF}}\right] &= F^2 \left(1-2\Psi_1^v\right)\!+\!\frac{1}{K}\left(F^2\Psi_2^v+ \tau^2\Psi_3^v \right) +\left(1 - \frac{1}{K}\right) F^2\Psi_2^d,
  \label{eq:divide-error}
\end{align}
where the effective number of data points which enters this formula is $N_\mathrm{eff}\!=\!N/K$ due to the splitting of the training set.
\begin{figure}[htb]
\centering
		\includegraphics[width=0.8\linewidth]{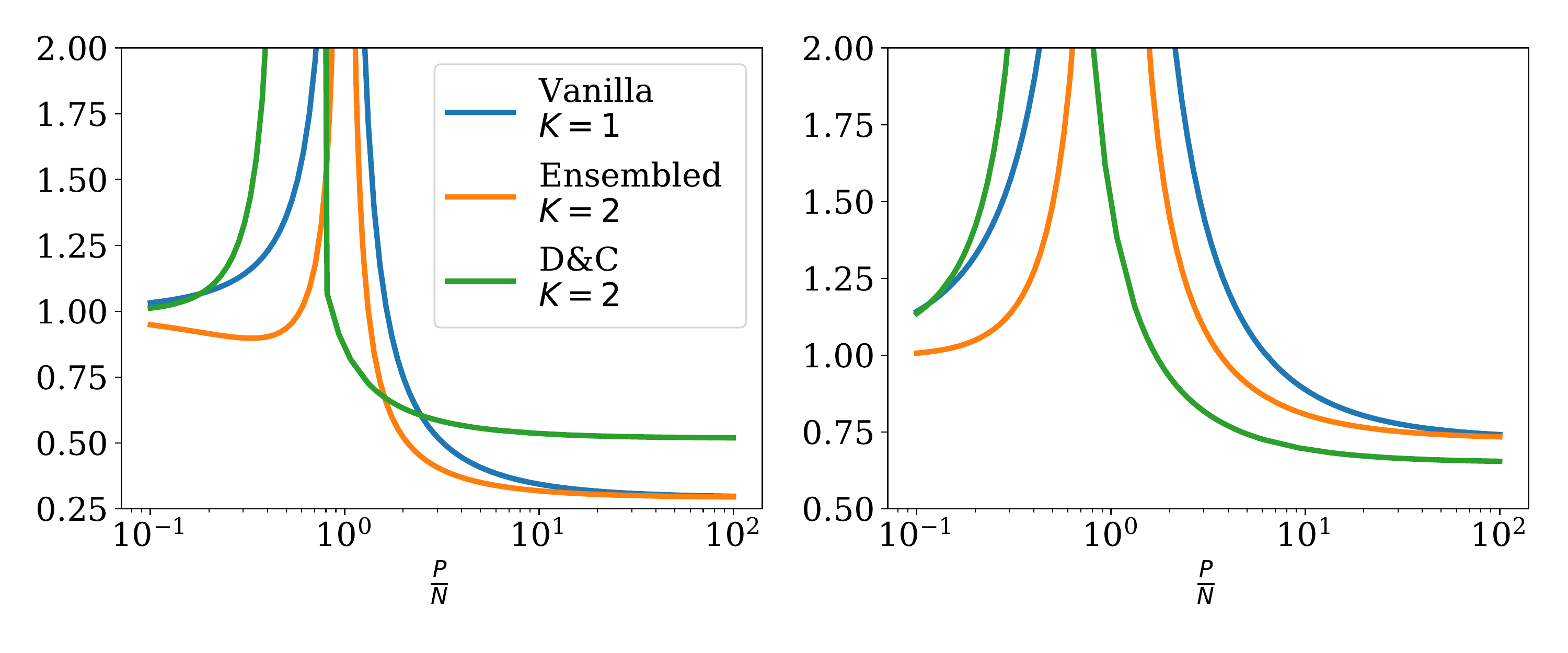}
		\vspace{-0.8cm}
			\caption{Comparison between performances of ensembling and divide and conquer for $K=2$ at different SNR. \textbf{Left:} SNR=$\frac{F}{\tau}=10$. \textbf{Right:} SNR=$\frac{F}{\tau}=1$ Computations performed with fixed $N/D=1$, $F=1$ and $\lambda=10^{-5}$.}
   \label{fig:divide_and_conquer}
\end{figure}

Comparing the previous expression with that obtained for ensembling~\eqref{eq:ensemble-error-app} is instructive: here, increasing $K$ replaces the \emph{vanilla} terms $\Psi_2^v, \Psi_3^v$ by the \emph{divide and conquer} term $\Psi_2^d$. This shows that divide and conquer has a \emph{denoising} effect: at $K\to\infty$, the effect of the additive noise on the labels is completely suppressed. This was not the case for ensembling. The price to pay is that $N_\mathrm{eff}$ decreases, hence one is shifted to the underparametrized regime.

In Figure~\ref{fig:divide_and_conquer}, we see that the kernel limit error of the divide and conquer approach, i.e. the asymptotic value of the error at $P/N\to\infty$, is different from the usual kernel limit error, since the effective dataset is two times smaller at $K=2$. The denoising effect of the divide and conquer approach is illustrated by the fact that its kernel limit error is higher at high SNR, but lower at low SNR. This is of practical relevance, and is much related to the beneficial effect of \emph{bagging} in noisy dataset scenarios. The divide and conquer approach, which only differs from bagging by the fact that the different partitions of the dataset are disjoint, was shown to reach bagging-like performance in various setups such as decision trees and neural networks~\cite{chawla2003distributed}.

\subsection{Is it always better to be overparametrized ?}
\label{app:noise}

A common thought is that the double descent curve always reaches its minimum in the over-parametrized regime, leading to the idea that the corresponding model "cannot overfit". In this section, we show that this is not always the case. Three factors tend to shift the optimal generalization to the underparametrized regime: (i) increasing the numbers of learners from which we average the predictions, $K$, (ii) decreasing the signal-to-noise ratio (SNR), $F/\tau$, and (iii) decreasing the size of the dataset, $N/D$. In other words, when ensembling on a small, noisy dataset, one is better off using an underparametrized model.

\begin{figure}[htb]
\centering
		\includegraphics[width=\linewidth]{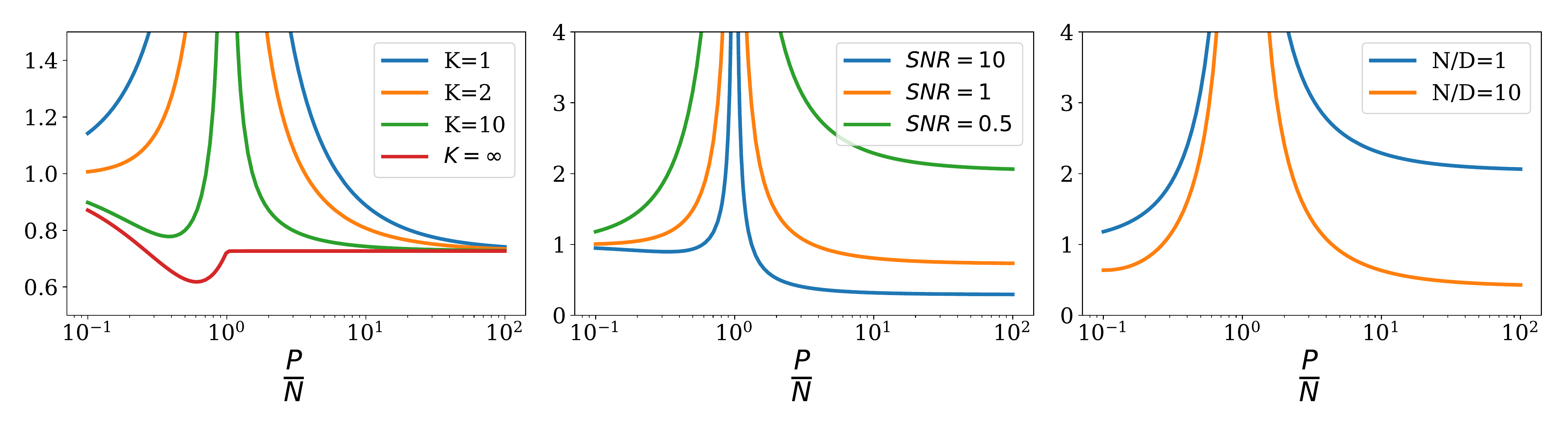}
			\caption{
			Generalisation error as a function of $P/N$: depending on the values of $K$, $F/\tau$ and $N/D$, optimal generalization can be reached in the underparametrized regime or the overparametrized regime. \textbf{Left:} $F/\tau=1$, $N/D=1$ and we vary $K$. This is the same as figure 5 in the main text, except that the higher noise causes the ensembling curve at $K\to\infty$ to exhibit a \emph{dip} in the underparametrized regime. \textbf{Center:} $K=2$, $N/D=1$ and we vary $F/\tau$. \textbf{Right:} $F/\tau=1$, $K=2$ and we vary $N/D$.}
   \label{fig:under_or_over}
\end{figure}

These three effects are shown in~\figref{fig:under_or_over}. In the left panel, we see that as we increase $K$, the minimum of generalization error jumps to the underparametrized regime $P<N$ for a high enough value of $K$. In the central/right panels, a similar effect occurs when decreasing the SNR or decreasing $N/D$.

\section{Statement of the Main Result}
\label{app:statement}

\subsection{Assumptions}
First, we state precisely the assumptions under which our main result is valid. Note, that these are the same as in \cite{mei2019generalization}.

\textbf{Assumption 1:}
$\sigma: \mathbb{R}\to\mathbb{R}$ is a weakly differential function with derivative $\sigma^\prime$. Assume there exists $c_0,c_1<\infty\in \sR$ such that for all $u\in\sR$ $|\sigma(u)|,|\sigma^\prime(u)|\leq c_0 \operatorname{e}^{c_1|u|}$. Then define:
\begin{equation}
    \mu_0=\E\left[\sigma(u)\right]\text{ } \text{ } \text{ }\mu_1=\E\left[u\sigma(u)\right]\text{ } \text{ }\mu_\star^2=\E\left[\sigma^2(u)\right] - \mu_0^2 - \mu_1^2,
    \label{eq:assumptionmu}
\end{equation}
where the expectation is over $u\sim\mathcal{N}(0,1)$. To facilitate readability, we specialize to the case $\mu_0=0$. This simply amounts to a shift ctivation function $\tilde{\sigma}$ of the network, $\tilde{\sigma}(x)=\sigma(x)-\mu_0$.

\textbf{Assumption 2:}
We work in the high-dimensional limit, i.e. in the limit where the input dimension $D$, the hidden layer dimension $P$ and the number of training points $N$ go to infinity with their ratios fixed. That is:
    \begin{equation}
        N,P,D\to\infty, \quad \text{  }\frac{P}{D}\equiv\psi_1=\mathcal{O}(1),\text{  }\text{  }\frac{N}{D}\equiv\psi_2=\mathcal{O}(1).
    \end{equation}
This condition implies that, in the computation of the risk $\mathcal{R}$, we can neglect all the terms of order $\mathcal{O}(1)$ in favour of the terms of order $\mathcal{O}(D)$. 

\textbf{Assumption 3:}
The labels are given by a linear ground truth, or teacher function:
\begin{align}
    y_\mu =&  f_d(\bs X_\mu) +\epsilon_\mu, \quad f_d(\bs x) = \langle\boldsymbol{\beta},\bs x\rangle,  \quad ||\boldsymbol{\beta}|| = F, \quad \epsilon_\mu\sim\mathcal{N}(0,\tau)\, .
\end{align}

Note that as explained in~\cite{mei2019generalization}, it is easy to add a non linear component to the teacher, but the latter would not be captured by the model (the student) in the regime $N/D=\mathcal{O}(1)$, and would simply amount to an extra noise term.

\subsection{Results}

Here we give the explicit form of the quantities appearing in our main result. In these expressions, the index $a\in\{v,e,d\}$ distinguishes the \emph{vanilla}, \emph{ensembling} and \emph{divide and conquer} terms.

\begin{align*}
\Psi_{1}&=\frac{1}{D}\operatorname{Tr}\left[H\left[S^v\right]^{-1} H\left[P_{\Psi_{1}}\right]\right],\\
\Psi_{2}^v&=\frac{1}{D}\operatorname{Tr}\left[H\left[S^v\right]^{-1}    H\left[P_{\Psi_{2}^v}\right]\right],\\
\Psi_{3}^v&=\frac{1}{D}\operatorname{Tr}\left[H\left[S^v\right]^{-1}    H\left[P_{\Psi_{3}^v}\right]\right],\\
\Psi_{2}^e&=\frac{1}{D}\operatorname{Tr}\left[H\left[S^e\right]^{-1}    H\left[P_{\Psi_{2}^e}\right]\right],\\
\Psi_{3}^e&=\frac{1}{D}\operatorname{Tr}\left[H\left[S^e\right]^{-1}    H\left[P_{\Psi_{3}^e}\right]\right],\\
\Psi_{2}^d&=\frac{1}{D}\operatorname{Tr}\left[H\left[S^d\right]^{-1}  H\left[P_{\Psi_{2}^d}\right]\right],\\
\end{align*}
where the Hessian matrix $H[F]$, for a given function $F:(q,r,\tilde q,\tilde r)\mapsto \mathbb R$ is defined as:
\begin{equation*}
    H\left[F\right]=\at{\begin{bmatrix}
    \frac{\partial F}{\partial q\partial q} &\frac{\partial F}{\partial q\partial r}&\frac{\partial F}{\partial q\partial \tilde{q}}&\frac{\partial F}{\partial q\partial\tilde{r}} \\
    \frac{\partial F}{\partial q\partial r}&\frac{\partial F}{\partial r\partial r} &\frac{\partial F}{\partial r\partial \tilde{q}} &\frac{\partial F}{\partial r\partial\tilde{r}} \\
    \frac{\partial F}{\partial q\partial \tilde{q}} &\frac{\partial F}{\partial r\partial \tilde{q}} &\frac{\partial F}{\partial \tilde{q}\partial \tilde{q}} &\frac{\partial F}{\partial \tilde{q}\partial\tilde{r}} \\
    \frac{\partial F}{\partial q\partial \tilde{r}} &\frac{\partial F}{\partial r\partial \tilde{r}}&\frac{\partial F}{\partial \tilde{q}\partial \tilde{r}}&\frac{\partial F}{\partial \tilde{r}\partial\tilde{r}}
    \end{bmatrix}}{\substack{q=q^*\\r=r^*\\\tilde r=0 \\\tilde q=0\\\\}},
\end{equation*}
with $q^\star$ and $r^\star$ being the solutions of the fixed point equation for the function $S_0:(q,r)\mapsto \mathbb R$ defined below:
\begin{align*}
    &\begin{cases}
    \frac{\partial S_0(q,r)}{\partial q}=0\\
    \frac{\partial S_0(q,r)}{\partial r}=0.
    \end{cases}\\
    &S_0(q,r)=\lambda \psi_1^2 \psi_2   q+\psi_2 \log \left(\frac{\mu_\star^2  \psi_1q}{\mu_1^2  \psi_1r+1}+1\right)+\frac{r}{q}+(1-\psi_1) \log (q)+\psi_2 \log \left(\mu_1^2  \psi_1 r+1\right)-\log (r).
\end{align*}

The explicit expression of the above quantities in terms of $(q,r,\tilde q, \tilde r)$ is given below.

\subsection{Explicit expression of $S^v, S^e, S^d$}
Here we present the explicit formulas for $S^v, S^e, S^d$,
which are defined as the functions $(q,r,\tilde q,\tilde r)\mapsto \mathbb R$ such that:
\begin{align*}
        S^v(q,r, \tilde q, \tilde r) &= 2 \left(S_0( q,r)+\tilde q f^v(q,r)  + \tilde r g^v(q,r)\right)\\
        S^e(q,r,\tilde{q},\tilde r) &= S_0(q,r) +\Tilde{r}^2 f^e(q,r) + \Tilde{q}^2 g^e(q,r)\\ 
        S^d(q,r,\Tilde{q},\Tilde{r})&= S_0(q,r)+ \tilde r^2 f^d(q,r)+ \tilde q^2 g^d(q,r),\\ 
\end{align*}
where we defined the functions $(q,r)\mapsto\mathbb R$,
\begin{align*}
        f^v(q,r)&=\lambda \psi_1^2 \psi_2+\frac{\mu_\star^2 \psi_1 \psi_2}{\mu_\star^2 \psi_1 q+\mu_1^2 \psi_1 r+1}+\frac{1-\psi_1}{q}-\frac{r}{q^2},\\       g^v(q,r)&=-\frac{\mu_\star^2 \mu_1^2 \psi_1^2 \psi_2 q}{\left(\mu_1^2 \psi_1 r+1\right) \left(\mu_\star^2 \psi_1 q+\mu_1^2 \psi_1 r+1\right)}+\frac{\mu_1^2 \psi_1 \psi_2}{\mu_1^2 \psi_1 r+1}+\frac{1}{q}-\frac{1}{r},\\
        f^e(q,r)&=\frac{2 r \mu_1^{2} \psi_1 \left( 1+q \mu_\star^2\psi_1\right)+\left(1+q \mu_\star^2\psi_1\right)^{2}-r^{2} \mu_1^{4} \psi_1^{2}(-1+\psi_2)}{r^{2}\left(1+r \mu_1^{2} \psi_1+q \mu_\star^2\psi_1\right)^{2}},\\
        g^e(q,r)&=\frac{\psi_1}{q^2},\\
        f^d(q,r)&=\frac{1}{r^{2}},\\
        g^d(q,r)&=\frac{\psi_1}{q^2}.
\end{align*}

\subsection{Explicit expression of $P_{\Psi_{1}}, P_{\Psi_{2}}^v, P_{\Psi_{3}}^v, P_{\Psi_{2}}^e, P_{\Psi_{3}}^e, P_{\Psi_{2}}^d$}

Here we present the explicit formulas for $P_{\Psi_{1}}, P_{\Psi_{2}}^v, P_{\Psi_{3}}^v, P_{\Psi_{2}}^e, P_{\Psi_{3}}^e, P_{\Psi_{2}}^d$,
which are defined as the functions $(q,r,\tilde q,\tilde r)\mapsto \mathbb R$ such that:
\begin{align*}
P_{\Psi_{1}}&=\psi_1 \psi_2 \mu_1^2 \left(M_X^{11}+\mu_1^2 \mu_\star^2 \psi_1^2 \left(M_X M_W^v M_X\right)^{11}+\mu_\star^2  \psi_1 \left(M_XM_W^v\right)^{11}\right),\\
P_{\Psi_{2}^v}&= D \psi_1^2 \psi_2  (\mu_1^2 \tilde r + \mu_\star^2 \tilde q) \left[ \mu_1^2 P_{XX}^v - 2 \mu_1^2 \mu_\star^2 \psi_1 P_{WX}^v + \mu_\star^2 P_{WW}^v \right],\\
P_{\Psi_{3}^v} &= D \psi_1^2 \psi_2  (\mu_1^2 \tilde r + \mu_\star^2 \tilde q) \left[\mu_1^2 \left( M_X^{12} + \mu_1^2 \mu_\star^2 \psi_1^2 \left[M_X M_W^v M_X\right]^{12}\right) - 2 \mu_1^2\mu_\star^2 \psi_1 \left[M_X M_W^v\right]^{12} + \mu_\star^2  \left[M_W^v\right]^{12} \right],\\
P_{\Psi_{2}^e}&=D \psi_1^2 \psi_2  \mu_1^2 \tilde r\left[\mu_1^2 P^e_{XX}-2 \mu_1^2 \mu_\star^2 \psi_1 P^e_{WX} + \mu_\star^2 P^e_{WW}\right],\\
P_{\Psi_{3}^e} &= D \psi_1^2 \psi_2 \mu_1^2  \tilde r \left[\mu_1^2 \left(M_X^{12} + \mu_1^2 \mu_\star^2 \psi_1^2 \left[M_X M_W^e M_X\right]^{12}\right) - 2 \mu_1^2\mu_\star^2 \psi_1 \left[M_X M_W^e\right]^{12} + \mu_\star^2  [M_W^e]^{12}  \right],\\
P_{\Psi_2^d} &= D \mu_1^2  \psi_1 \psi_2^2  \tilde{r} \left[ \psi_1 \mu_1^2 P_{XX} + 2 \mu_\star^2 \mu_1^2 \psi_1^2 P_{WX} + \mu_\star^2 \psi_1 P_{WW} \right],\\
\end{align*}
where we defined the scalars $P_{XX}, P_{WX}, P_{WW}$ as follows:
\begin{align*}
P^{v}_{XX} &= \psi_2 N_X^{12} +M_X^{12}+ 2 \psi_2(\mu_1 \mu_\star\psi_1)^2 \left[M_X N_X M_W^{a}\right]^{12} +(\mu_1 \mu_\star\psi_1)^2\left[M_X M_W^{a} M_X\right]^{12} \\
&+ \psi_2(\mu_1 \mu_\star\psi_1)^4 \left[M_X M_W N_X M_W^{a} M_X\right]^{12},\\
P^{v}_{WX} &= \psi_2 \left[N_X M_W^{a}\right]^{12} + \left[M_X M_W^{a}\right]^{12}+\psi_2 (\mu_1 \mu_\star\psi_1)^2 \left[M_X M_W^{a} N_X M_W^{a}\right]^{12}, \\
P^{v}_{WW} &= [M_W^{a}]^{12}+ \psi_2(\mu_1 \mu_\star\psi_1)^2 \left[M_W^{a} N_X M_W^{a}\right]^{12} ,\\
P^{e}_{XX} &= P^{v}_{XX},\\
P^{e}_{WX} &= P^{v}_{WX},\\
P^{e}_{WW} &= P^{v}_{WW},\\
P^d_{XX} &= \left( N_X^{d11} + 2(\mu_1 \mu_\star\psi_1)^2 \left[N^d_X M^d_W M^d_X\right]^{11} + (\mu_1 \mu_\star\psi_1)^4 \left[M^d_X M^d_W N^d_X M^d_W M^d_X\right]^{11}\right),\\
P^d_{WX} &=  \left[N^d_X M^d_W\right]^{11} + (\mu_1 \mu_\star\psi_1)^2 \left[M^d_X M^d_W N^d_X M^d_W\right]^{11},  \\
P^d_{WW} &=(\mu_1 \mu_\star\psi_1)^2 \left[M^d_W N^d_X M^d_W\right]^{11},
\end{align*}
and the $2\times2$ matrices $M_X, M_W, N_X$ as follows:
\begin{align*}
&M_X^v =\begin{bmatrix}
        \frac{r}{1+\mu_1^2\psi_1 r}&\frac{\tilde{r}}{\left(1+\mu_1^2\psi_1 r\right)^2}\\
        \frac{\tilde{r}}{\left(1+\mu_1^2\psi_1 r\right)^2}&\frac{r}{1+\mu_1^2\psi_1 r}
    \end{bmatrix}, \quad
M^v_W=\begin{bmatrix}
       \frac{q\left(1+\mu_1^2\psi_1 r\right)}{1+2\mu_1^2\psi_1 r+\mu_\star^2\psi_1 q}&
       \frac{q^2 \mu_1^2\mu_\star^2\psi^2_1\tilde{r}}{\left(1+\mu_1^2\psi_1 r+\mu_\star^2\psi_1 q\right)^2}\\
       \frac{q^2 \mu_1^2\mu_\star^2\psi^2_1\tilde{r}}{\left(1+\mu_1^2\psi_1 r+\mu_\star^2\psi_1 q\right)^2} &\frac{q\left(1+\mu_1^2\psi_1 r\right)}{1+2\mu_1^2\psi_1 r+\mu_\star^2\psi_1 q}
    \end{bmatrix}, \quad 
N^v_{X} =\frac{1}{\left(1+\mu_1^2\psi_1 r\right)^2}\begin{bmatrix}
       r&\tilde{r}\\
      \tilde{r}&r
    \end{bmatrix},
\\
&M^e_X=M^v_X, \quad
M^e_W=M^v_W+\frac{(1 + r \mu_1^2 \psi_1)^2 \tilde q}{\left(1+\mu_1^2\psi_1 r+\mu_\star^2\psi_1 q\right)^2}\begin{bmatrix}
       0&1\\
     1&0
    \end{bmatrix}, \quad
N^e_{X} =\frac{1}{\left(1+\mu_1^2\psi_1 r\right)^2}\begin{bmatrix}
       r&\tilde{r}\\
      \tilde{r}&r
    \end{bmatrix},
\\
&M_X^d= \frac{r}{1 + \mu_1^2 \psi_1 r^2}\begin{bmatrix}
      1&0\\
      0&1
    \end{bmatrix},\quad
M_W^d= \frac{q (1 + \mu_1^2 \psi_1 r)}{1+\mu_1^2 \psi_1 r+\mu_\star^2 \psi_1 q}\begin{bmatrix}
      1&0\\
      0&1
    \end{bmatrix},\quad
N_X^d=\frac{1}{(1+\mu_1^2 \psi_1 r)^2}\begin{bmatrix}
       \tilde r&0\\
      0&\tilde r
    \end{bmatrix}.
\end{align*}

\section{Replica Computation}
\label{app:computation}

\subsection{Toolkit}

\subsubsection{Gaussian integrals}
In order to obtain the main result for the generalisation error, we perform the averages over all the sources of randomness in the system in the following order: over the dataset $X$, then over the noise $W$, and finally over the random feature layers $\Theta$. Here are some useful formulaes used throughout the computations:
\begin{align}
\begin{cases}
    \int e^{-\frac{1}{2} x_i G_{ij} x_j + J_i x_i} dx &= (\det G)^{-\frac{1}{2}} e^{\frac{1}{2} J_i G^{-1}_{ij} J_j}, \\
    \int x_a e^{-\frac{1}{2} x_i G_{ij} x_j + J_i x_i} dx &= P^1_a (\det G)^{-\frac{1}{2}} e^{\frac{1}{2} J_i G^{-1}_{ij} J_j}, \\
    \int x_a x_b e^{-\frac{1}{2} x_i G_{ij} x_j + J_i x_i} dx &=  P^2_{ab} (\det G)^{-\frac{1}{2}} e^{\frac{1}{2} J_i G^{-1}_{ij} J_j},\\
    \int x_a x_b x_c e^{-\frac{1}{2} x_i G_{ij} x_j + J_i x_i} dx &=  P^3_{abc} (\det G)^{-\frac{1}{2}} e^{\frac{1}{2} J_i G^{-1}_{ij} J_j},\\
    \int x_a x_b x_c x_d e^{-\frac{1}{2} x_i G_{ij} x_j + J_i x_i} dx &= P^4_{abcd} (\det G)^{-\frac{1}{2}} e^{\frac{1}{2} J_i G^{-1}_{ij} J_j},
\end{cases}
\label{eq:integrals}
\end{align} 

with
\begin{align*}
    &P^1_{a} = [G^{-1}J]_a , \\
    &P^2_{ab} = ((G^{-1})_{ab} + [G^{-1}J]_a [G^{-1}J]_b) , \\
    &P^3_{abc} = \sum_{\overline a, \overline b, \overline c \in perm(abc)} \left( (G^{-1})_{\overline a \overline b} [G^{-1}J]_{\overline{c}} + [G^{-1}J]_{\overline{a}} [G^{-1}J]_{\overline{b}} [G^{-1}J]_{\overline{c}} \right) ,\\
    &P^4_{abcd} = \sum_{\overline a, \overline b, \overline c, \overline d \in perm(abcd)} \left( (G^{-1})_{\overline a \overline b} (G^{-1})_{\overline c \overline d} + [G^{-1}J]_{\overline{a}} [G^{-1}J]_{\overline{b}} [G^{-1}J]_{\overline{c}} [G^{-1}J]_{\overline{d}} + (G^{-1})_{\overline a \overline b} [G^{-1}J]_{\overline{c}} [G^{-1}J]_{\overline{d})} \right).
\end{align*}

\subsubsection{Replica representation of an inverse matrix}

To obtain gaussian integrals we will use the "replica" representation the element $(ij)$ of a matrix $M$ of size $D$:
\begin{equation}
M^{-1}_{ij} = \lim_{n\to 0} \int \left(\prod_{\alpha=1}^{n}\prod_{i=1}^{D} d\eta_i^\alpha\right) \eta^{1}_i \eta^{1}_j \exp\left(-\frac{1}{2} \eta_i^\alpha M_{ij} \eta_j^\alpha\right).
\label{eq:replica-inverse}
\end{equation}

Indeed, using the gaussian integral representation of the inverse of $M$,
\begin{align*}
M^{-1}_{ij} &= \mathcal{\bs Z}^{-1} 
\int \left(\prod_{i=1}^{D} d\eta_i\right) \eta_i \eta_j \exp\left(-\frac{1}{2} \eta_i M_{ij} \eta_j\right) ,\\
\mathcal{Z} &= \sqrt{\frac{(2\pi)^D}{\det M}}\\
&=\int \left(\prod_{i=1}^{D} d\eta_i\right) \exp\left(-\frac{1}{2} \eta_i M_{ij} \eta_j\right).
\end{align*}

Using the replica identity, we rewrite this as 
\begin{align*}
M^{-1}_{ij} &= \lim_{n\to 0} \mathcal{Z}^{n-1} 
\int \left(\prod_{i=1}^{D} d\eta_i\right) \eta_i \eta_j \exp\left(-\frac{1}{2} \eta_i M_{ij} \eta_j\right).
\end{align*}

Renaming the integration variable of the integral on the left as $\eta^1$ and the $n-1$ others as $\eta^\alpha, \alpha\in\{2,n\}$, we obtain expression~\eqref{eq:replica-inverse}.



\subsection{The Random Feature model}

In what follows, we will explicitly leave the indices of all the quantities used. We use the notation, called Einstein summation convention in physics, in which all repeated indices are summed but the sum is not explicitly written. Indices $i\in\{1...D\}$ are used to refer to the input dimension, $h\in\{1...P\}$ to refer to the hidden layer dimension and $\mu\in\{1...N\}$ to refer to the number of data points. 

\subsubsection{With a single learner}

In the random features model, the predictor can be computed explicitly:
\begin{align}
    \hat a &= \frac{1}{\sqrt D}\bs{y}^{\mathrm{T}}_\mu \bs{Z}_{\mu h}\left(\bs{Z}^{\mathrm{T}} \bs{Z}+\psi_{1} \psi_{2} \lambda \mathbf{I}_{N}\right)^{-1}_{h h'} \\
    f(\bs x) &= \hat a_h \sigma\left(\frac{\bs\Theta_{h'i} \bs{x}_i}{ \sqrt{D}}\right) \\ 
    &=\bs{y}^{\mathrm{T}}_\mu \bs{Z}_{\mu h}\left(\bs{Z}^{\mathrm{T}} \bs{Z}+\psi_{1} \psi_{2} \lambda \mathbf{I}_{N}\right)^{-1}_{h h'} \sigma\left(\frac{\bs\Theta_{h'i} \bs{x}_i}{ \sqrt{D}}\right) / \sqrt{D},
\end{align}
where 
\begin{align}
    \bs{y}_\mu &=f_{d}\left(\bs{X}_{\mu}\right) + \bs \epsilon_\mu, \\ 
    \bs{Z}_{\mu h}&=\frac{1}{\sqrt{D}} \sigma\left(\frac{1}{\sqrt{D}} \bs{\Theta}_{hi}\bs{X}_{\mu i}\right).
\end{align}

Hence the generalization error can be computed as:
\begin{align} 
\mathcal{R}_{\mathrm{RF}}=& \mathbb{E}_{\bs{x}}\left[\left(f_{d}(\bs{x})-\bs{y}^{\mathrm{T}}_\mu \bs{Z}_{\mu h}\left(\bs{Z}^{\mathrm{T}} \bs{Z}+\psi_{1} \psi_{2} \lambda \mathbf{I}_{N}\right)^{-1}_{h h'} \sigma\left(\frac{\bs\Theta_{h'i} \bs{x}_i}{ \sqrt{D}}\right) / \sqrt{D}\right)^{2}\right] \\ 
=& \mathbb{E}_{\bs{x}}\left[f_{d}(\bs{x})^{2}\right]-2 \bs{y}_\mu^{\top} \bs{Z}_{\mu h}\left(\bs{Z}^{\mathrm{T}} \bs{Z}+\psi_{1} \psi_{2} \lambda \mathbf{I}_{N}\right)^{-1}_{hh'} \bs{V}_{h'} / \sqrt{D} \notag \\ &+\bs{y}_\mu^{\mathrm{T}} \bs{Z}_{\mu h}\left(\bs{Z}^{\mathrm{T}} \bs{Z}+\psi_{1} \psi_{2} \lambda \mathbf{I}_{N}\right)^{-1}_{hh_1} \bs{U}_{h_1 h_2}\left(\bs{Z}^{\mathrm{T}} \bs{Z}+\psi_{1} \psi_{2} \lambda \mathbf{I}_{N}\right)^{-1}_{h_2 h'} \bs{Z}_{h'\mu'}^{\mathrm{T}} \bs{y}_{\mu'} / D ,
\end{align}
where 
\begin{align} 
\bs V_{h} &=\mathbb{E}_{\bs{x}}\left[f_{d}(\bs{x}) \sigma\left(\frac{\langle\bs{\Theta}_{hi} \bs{x}_i\rangle}{\sqrt{D}}\right)\right] ,\\ 
\bs U_{h h'} &=\mathbb{E}_{\bs{x}}\left[\sigma\left(\frac{\langle\bs{\Theta}_{hi} \bs{x}_i \rangle}{\sqrt{D}}\right) \sigma\left(\frac{\bs{\Theta}_{h'i}, \bs{x}_i}{\sqrt{D}} \right)\right].
\label{eq:u_v_1}
\end{align}

\subsubsection{Ensembling over $K$ learners}

When ensembling over $K$ learners with independently sampled random feature vectors, the predictor becomes:
\begin{equation}
    f(\bs x) =\frac{1}{K \sqrt{D}} \sum_{k}\bs{y}^{\mathrm{T}}_\mu \bs{Z}^{(k)}_{\mu h}\left(\bs{Z}^{\mathrm{T}(k)} \bs{Z}^{(k)}+\psi_{1} \psi_{2} \lambda \mathbf{I}_{N}\right)^{-1}_{h h'} \sigma\left(\frac{\bs\Theta^{(k)}_{h'i} \bs{x}_i}{ \sqrt{D}}\right),
\end{equation}
where 
\begin{align}
    \bs{Z}^{(k)}_{\mu h}&=\frac{1}{\sqrt{D}} \sigma\left(\frac{1}{\sqrt{D}} \bs{\Theta}^{(k)}_{hi}\bs{X}_{\mu i}\right).
\end{align}

The generalisation error is then given by:
\begin{align} 
\mathcal{R}_{\mathrm{RF}}=& \mathbb{E}_{\bs{x}}\left[\left(f_{d}(\bs{x})-\frac{1}{K}\sum_k\bs{y}^{\mathrm{T}} \bs{Z}^{(k)}\left(\bs{Z}^{\mathrm{T}(k)} \bs{Z}^{(k)}+\psi_{1} \psi_{2} \lambda \mathbf{I}_{N}\right)^{-1} \sigma\left(\frac{\bs\Theta^{(k)}_{h'i} \bs{x}_i}{ \sqrt{D}}\right) / \sqrt{D}\right)^{2}\right] \\
=& \mathbb{E}_{\bs{x}}\left[f_{d}(\bs{x})^{2}\right]-\frac{2}{K}\sum_k \bs{y}^{\top} \bs{Z}^{(k)}\left(\bs{Z}^{\mathrm{T}(k)} \bs{Z}^{(k)}+\psi_{1} \psi_{2} \lambda \mathbf{I}_{N}\right)^{-1} \bs{V}^{(k)} / \sqrt{D} \notag \\ &+\frac{1}{K^2}\sum_{\substack{k\\l\neq k}}\bs{y}^{\mathrm{T}} \bs{Z}^{(k)}\left(\bs{Z}^{\mathrm{T}(k)} \bs{Z}^{(k)}+\psi_{1} \psi_{2} \lambda \mathbf{I}_{N}\right)^{-1} \bs{U}^{(kl)}\left(\bs{Z}^{\mathrm{T}(l)} \bs{Z}^{(l)}+\psi_{1} \psi_{2} \lambda \mathbf{I}_{N}\right)^{-1} \bs{Z^{(l)}\mathrm{T}} \bs{y} / D,
\end{align}
where
\begin{align}
\bs V^{(k)}_{h}&=\mathbb{E}_{\bs{x}}\left[f_{d}(\bs{x}) \sigma\left(\frac{\langle\bs{\Theta}^{(k)}_{hi} \bs{x}_i\rangle}{\sqrt{D}}\right)\right] ,\\
\bs U^{(kl)}_{h h'} &=\mathbb{E}_{\bs{x}}\left[\sigma\left(\frac{\langle\bs{\Theta}^{(k)}_{hi} \bs{x}_i\rangle}{ \sqrt{D}}\right) \sigma\left(\frac{\langle\bs{\Theta}^{(l)}_{h'i'} \bs{x}_{i'}\rangle}{\sqrt{D}}\right)\right].
\end{align}

\subsubsection{Equivalent Gaussian Covariate Model}

It was shown in \cite{mei2019generalization} that the random features model is equivalent, in the high-dimensional limit of Assumption 2, to a Gaussian covariate model in which the activation function $\sigma$ is replaced as:
\begin{equation}
     \sigma\left(\frac{ \bs \Theta_{hi}^{(k)} \bs X_{\mu i}}{\sqrt{D}}\right) \to \mu_0 + \mu_1\frac{ \bs \Theta_{hi}^{(k)} \bs X_{\mu i}}{\sqrt{D} }+ \mu_\star \bs W^{(k)}_{\mu h},
\end{equation}
with $\bs W^{(k)}\in \mathbb{R}^{N\times P}$, $W^{(k)}_{\mu h} \sim \mathcal{N}(0,1)$ and $\mu_0$, $\mu_1$ and $\mu_\star$ defined in \eqref{eq:assumptionmu}. To simplify the calculations, we take $\mu_0=0$, which amounts to adding a constant term to the activation function $\sigma$.

This powerful mapping allows to express the quantities $\bs U, \bs V$. We will not repeat their calculations here: the only difference here is $\bs U^{kl}$, which carries extra indices $k,l$ due to the different initialization of the random features $\bs \Theta^{(k)}$. In our case, 
\begin{align}
    \boldsymbol{U}^{(kl)}_{hh'} &= \frac{\mu_{1}^{2}}{D}\boldsymbol{\Theta}^{(k)}_{hi}\boldsymbol{\Theta}_{h'i}^{(l)}+\mu_{\star}^{2}\delta_{kl}\delta_{hh'}.
\end{align}

Hence we can rewrite the generalization error as
\begin{align}
   \E_{\{\bs \Theta^{(k)}\}, \bs X, \boldsymbol{\varepsilon}} \left[ \mathcal{R}_{\mathrm{RF}}\right]=F^2 \left(1-2\Psi_1^v\right)+\frac{1}{K}\left(F^2\Psi_2^v+ \tau^2\Psi_3^v \right)
   +\left(1-\frac{1}{K}\right) \left(F^2\Psi_2^e + \tau^2\Psi_3^e \right),
   \label{eq:ensemble-error-app}
\end{align}
where $\Psi_1$,$\Psi_2^v$,$\Psi_2^e$,$\Psi_3^v$,$\Psi_3^e$ are given by:
\begin{align*}
    \Psi_{1}&=\frac{1}{D} \operatorname{Tr}\left[\left(\frac{\mu_{1}}{D} \boldsymbol{X} \bs\Theta^{(1)\top}\right)^{\top} \boldsymbol{Z^{(1)}}\left(\boldsymbol{Z^{(1)\top}} \boldsymbol{Z^{(1)}}+\psi_{1} \psi_{2} \lambda \mathbf{I}_{N}\right)^{-1}\right],\\
    \Psi_{2}^v&=\frac{1}{D} \operatorname{Tr}\left[\left(\boldsymbol{Z^{(1)\top}} \boldsymbol{Z^{(1)}}+\psi_{1} \psi_{2} \lambda \mathbf{I}_{N}\right)^{-1}\left( \frac{\mu_{1}^{2}}{D} \bs\Theta^{(1)} \bs\Theta^{(1)\top}+\mu_{*}^{2} \mathbf{I}_{N}\right)\left(\boldsymbol{Z^{(1)\top}} \boldsymbol{Z^{(1)}}+\psi_{1} \psi_{2} \lambda \mathbf{I}_{N}\right)^{-1} \boldsymbol{Z^{(1)\top}} \left(\frac{1}{D} \boldsymbol{X} \boldsymbol{X}^{\top}\right) \boldsymbol{Z^{(1)}}\right],\\
     \Psi_{3}^v&=\frac{1}{D} \operatorname{Tr}\left[\left(\boldsymbol{Z^{(1)\top}} \boldsymbol{Z^{(1)}}+\psi_{1} \psi_{2} \lambda \mathbf{I}_{N}\right)^{-1}\left( \frac{\mu_{1}^{2}}{D} \bs\Theta^{(1)} \bs\Theta^{(1)\top}+\mu_{\star}^{2} \mathbf{I}_{N}\right)\left(\boldsymbol{Z^{(1)\top}} \boldsymbol{Z^{(1)}}+\psi_{1} \psi_{2} \lambda \mathbf{I}_{N}\right)^{-1} \boldsymbol{Z^{(1)\top}} \boldsymbol{Z^{(1)}}\right],\\
     \Psi_{2}^e&=\frac{1}{D} \operatorname{Tr}\left[\left(\boldsymbol{Z^{(1)\top}} \boldsymbol{Z^{(1)}}+\psi_{1} \psi_{2} \lambda \mathbf{I}_{N}\right)^{-1}\left( \frac{\mu_{1}^{2}}{D} \bs\Theta^{(1)} \bs\Theta^{(2)\top}\right)\left(\boldsymbol{Z^{(2)\top}} \boldsymbol{Z^{(2)}}+\psi_{1} \psi_{2} \lambda \mathbf{I}_{N}\right)^{-1} \boldsymbol{Z^{(2)\top}} \left(\frac{1}{D} \boldsymbol{X} \boldsymbol{X}^{\top}\right) \boldsymbol{Z^{(1)}}\right],\\
     \Psi_{3}^e&=\frac{1}{D} \operatorname{Tr}\left[\left(\boldsymbol{Z^{(1)\top}} \boldsymbol{Z^{(1)}}+\psi_{1} \psi_{2} \lambda \mathbf{I}_{N}\right)^{-1}\left( \frac{\mu_{1}^{2}}{D} \bs\Theta^{(1)} \bs\Theta^{(2)\top}\right)\left(\boldsymbol{Z^{(2)\top}} \boldsymbol{Z^{(2)}}+\psi_{1} \psi_{2} \lambda \mathbf{I}_{N}\right)^{-1} \boldsymbol{Z^{(2)\top}} \boldsymbol{Z^{(1)}}\right].
\end{align*}

\subsection{Computation of the \emph{vanilla} terms}
To start with, let us compute the vanilla terms (those who carry a superscript $v$), which involve a single instance of the random feature vectors. Note that these were calculated in~\cite{mei2019generalization} by evaluating the Stieljes transform of the random matrices of which we need to calculate the trace. The replica method used here makes the calculation of the vanilla terms carry over easily to the the ensembling terms (superscript $e$) and the divide and conquer term (superscript $d$). To illustrate the calculation steps, we will calculate $\Psi_3^v$, then provide the results for $\Psi_2^v$ and $\Psi_1$.

In the vanilla terms, the two inverse matrices that appear are the same. Hence we use twice the replica identity~\eqref{eq:replica-inverse}, introducing $2n$ replicas which all play the same role:
\begin{align}
M^{-1}_{ij} M^{-1}_{kl} = \lim_{n\to 0} \int \left(\prod_{\alpha=1}^{2n} d\eta\right) \eta^{1}_i \eta^{1}_j \eta^{2}_k \eta^{2}_l \exp\left(-\frac{1}{2} \eta^\alpha M_{ij} \eta^\alpha\right).
\label{eq:replica_formula_vanila}
\end{align}

The first step is to perform the averages, i.e. the Gaussian integrals, over the dataset $X$, the deterministic noise $W$ induced by the non-linearity of the activation function and the random features $\Theta$.

\subsubsection{Averaging over the dataset}
Replacing the activation function by its Gaussian covariate equivalent model and using \eqref{eq:replica_formula_vanila}, the term $\Psi_3$ can be expanded as:
\begin{align*}
\begin{split}
    \Psi_3^v &= \frac{1}{D} \left[ \bs Z_{\mu h}\left({\bs Z}^{\top} {\bs Z} + \psi_{1} \psi_{2} \lambda \mathbf{I}_{N}\right)_{h h_1}^{-1} \left(\frac{\mu_1^2}{D}\bs \Theta_{h_1 i} \bs \Theta_{h_2 i} + \mu_\star^2 \delta_{h_1 h_2}\right) \left({\bs Z}^{\top} {\bs Z}+\psi_{1} \psi_{2} \lambda \mathbf{I}_{N}\right)_{h_2 h'}^{-1} \bs Z_{h'\mu} \right]\\
    &= \frac{1}{D} \left(\frac{\mu_1^2}{D}\bs \Theta_{h_1 i} \bs \Theta_{h_2 i} + \mu_\star^2 \delta_{h_1 h_2}\right)  \left[ \bs Z_{\mu h} \bs Z_{h'\mu} \right] \int \left(\prod_{\alpha=1}^{2n} d\eta^\alpha \right) \eta^{1}_h \eta^{1}_{h_1} \eta^{2}_{h_2} \eta^{2}_{h'} \exp\left(-\frac{1}{2} \eta^{\alpha}_h \left({\bs Z}^{\top} {\bs Z} + \psi_{1} \psi_{2} \lambda \mathbf{I}_{N}\right)_{h h'} \eta^{\alpha}_{h'} \right)\\
    &= \frac{1}{D^2} \left(\frac{\mu_1^2}{D}\bs \Theta_{h_1 i} \bs \Theta_{h_2 i} + \mu_\star^2 \delta_{h_1 h_2}\right)  \left( \frac{\mu_1}{\sqrt D} \bs \Theta \bs X + \mu_\star \bs W \right)_{h\mu} \left( \frac{\mu_1}{\sqrt D} \bs \Theta \bs X + \mu_\star \bs W \right)_{h'\mu} \\
    &\int \left(\prod_{\alpha=1}^{2n} d\eta^\alpha \right) \eta^{1}_h \eta^{1}_{h_1} \eta^{2}_{h_2} \eta^{2}_{h'} \exp\left(-\frac{1}{2} \eta^{\alpha}_h \left( \frac{1}{D}\left( \frac{\mu_1}{\sqrt D} \bs \Theta \bs X + \mu_\star \bs W \right)_{h\mu} \left( \frac{\mu_1}{\sqrt D} \bs \Theta \bs X + \mu_\star \bs W \right)_{h'\mu} + \psi_{1} \psi_{2} \lambda \delta_{hh'}\right) \eta^{\alpha}_{h'} \right).
    \end{split}
\end{align*}
Now, we introduce $\lambda_i^\alpha:= \frac{1}{\sqrt{P}}\eta_h^\alpha \bs \Theta_{hi}$, and enforce this relation using the Fourier representation of the delta-function:
\begin{align}
	1 = \int d\lambda^{\alpha }_i d\hat \lambda^{\alpha }_i e^{i \hat \lambda^\alpha_i (\sqrt P \lambda^\alpha_i - \eta^{\alpha }_h \bs \Theta_{hi})}.
\end{align}
The average over the dataset $\bs X_{\mu i}$ has the form of \eqref{eq:integrals} with:
\begin{align}
    (G_X)_{\mu \mu', ii'} &= \delta_{\mu \mu'}\left( \delta_{i i'} + \frac{\mu_1^2 \psi_1}{D} \lambda^{\alpha }_i \lambda^{\alpha }_{i'} \right),\\
    (J_X)_{\mu, i} &= \frac{\mu_1 \mu_\star \sqrt{\psi_1}}{D} \sum_{\alpha a} \lambda^{\alpha }_i \eta^{\alpha }_{h} \bs W_{\mu h}.
    \end{align}
Using formulae \eqref{eq:integrals}, we obtain:
\begin{align*}
\begin{split}
   \Psi_3^v = &\frac{N}{D^2} \int \left(\prod_{\alpha=1}^{2n} d\eta^\alpha d\lambda^\alpha d\hat{\lambda}^\alpha\right) \left(\frac{\mu_1^2 P}{D}\lambda^{1}_{i} \lambda^{2}_{i} + \mu_\star^2 \eta^{1}_{i} \eta^{2}_{i}\right) \\
   &\quad \left[ \mu_\star^2 \eta^{1}_h  \bs W_{h} \bs W_{h'} \eta^{2}_{h'} + \mu_1^2 \psi_1 \lambda^{1}_{i} \lambda^{2}_{i'} \left( (G^{-1}_X)_{i i'} + (G^{-1}_X J_X)_{i} (G^{-1}_X J_X)_{i'} \right) + 2 \mu_1\mu_\star \sqrt{\psi_1} \lambda_i^1 \bs W_{h} \eta_{h}^2 (G^{-1}_X J_X)_{i}\right] \\
    &\exp\!\left( \!-\frac{n}{2}\log \det(G_X) \!-\!\frac{1}{2} \eta^{\alpha }_h \left( \frac{\mu_\star^2}{D} \bs W_{h} \bs W_{h'} \delta_{\alpha \beta} \!-\! \frac{\mu_1^2 \mu_\star^2 \psi_1}{D^2} \bs W_{h} \lambda^{\alpha }_i (G_X)^{-1}_{ii'} \lambda^{\beta}_{i'} \bs W_{h'} \!+\! \psi_{1} \psi_{2} \lambda \delta_{hh'}\right) \eta^{\beta}_{h'}\!+\!i \hat \lambda^\alpha_i \!(\sqrt P \lambda^\alpha_i \!-\! \eta^{\alpha }_h \bs \Theta_{hi})\! \right)
\end{split}
\end{align*}

Note that due to with a slight abuse of notation we got rid of indices $\mu$, which all sum up trivially to give a global factor $N$.

\subsubsection{Averaging over the deterministic noise}

The expectation over the deterministic noise $\bs W_{h}$ is a Gausssian integral of the form \eqref{eq:integrals} with:
\begin{align}
    \left[G_{W}\right]_{hh'} &= \delta_{hh'} +\frac{\mu_\star^2}{D} \eta^{\alpha }_h A^{\alpha \beta} \eta^{\beta}_{h'} ,\\
    \left[J_W\right]_{h} &= 0,\\
    A^{\alpha \beta} &= \delta_{\alpha\beta} -\mu_1^2 \psi_1 \frac{1}{D} \sum_{i,j} \lambda_i ^{\alpha}[G_X^{-1}]_{ij} \lambda_j ^{\beta}.
\end{align}

Note that the prefactor involves, constant, linear and quadratic terms in $\bs W$ since:
\begin{align*}
    (G^{-1}_X J_X)_{i} = \frac{\mu_1 \mu_\star \sqrt{\psi_1}}{D} \left[\eta^{\alpha } \bs W\right] \left[ G^{-1}_X \lambda^{\alpha }\right]_i .
\end{align*}
Thus, one obtains:
\begin{align*}
\begin{split}
    \Psi_3^v = \frac{\psi_2}{D} &\int \left(\prod_{\alpha=1}^{2n} d\eta^\alpha d\lambda^\alpha d\hat{\lambda}^\alpha\right) \left(\mu_1^2 \psi_1\lambda^{1}_{i} \lambda^{2}_{i} + \mu_\star^2 \eta^{1}_{i} \eta^{2}_{i}\right) \left[ \mu_\star^2 \left[\eta^{1}(G^{-1}_W)\eta^{2} \right] + \mu_1^2 \psi_1 \left[\lambda^{1}  H_W \lambda^{2} \right] + 2 \mu_1^2\mu_\star^2 \psi_1 [\lambda^1 S_W \eta^2] \right] \\
    & \exp\left( -\frac{n}{2}\log \det(G_X) -\frac{n}{2}\log \det(G_W) -\frac{1}{2}  \psi_{1} \psi_{2} \lambda \sum (\eta^{\alpha }_h)^2+i \hat \lambda^\alpha_i (\sqrt P \lambda^\alpha_i - \eta^{\alpha }_h \bs \Theta_{hi}) \right),
    \end{split}
\end{align*}
with 
\begin{align}
    (H_W)_{ij} &= (G^{-1}_X)_{i j} + \frac{\mu_1^2 \mu_\star^2 \psi_1}{D^2} \left[\eta^{\alpha } (G^{-1}_W) \eta^{\beta}\right] \left[ G^{-1}_X \lambda^{\alpha }\right]_i \left[ G^{-1}_X \lambda^{\beta}\right]_{j},  \\
    (S_W)_{ih} &=\frac{1}{D} \left[ G^{-1}_X\lambda^\alpha\right]_i \left[ G_W^{-1}\eta^\alpha\right]_h .
\end{align}

\subsubsection{Averaging over the random feature vectors}
The expectation over the random feature vectors $\bs \Theta_{hi}$ is a Gausssian integral of the form \eqref{eq:integrals} with:
\begin{align}
    \left[G_\Theta\right]_{hh', ii'} &= \delta_{hh', ii'},\\
    \left[J_\Theta\right]_{hi} &= - i \hat \lambda^{\alpha }_i \eta^{\alpha }_h.
\end{align}
Preforming this integration results in:
\begin{align*}
\begin{split}
    \Psi_3^v = \frac{\psi_2}{D} &\int \left(\prod_{\alpha=1}^{2n} d\eta^\alpha d\lambda^\alpha d\hat{\lambda}^\alpha\right) \left(\mu_1^2 \psi_1\lambda^{1}_{i} \lambda^{2}_{i} + \mu_\star^2 \eta^{1}_{i} \eta^{2}_{i}\right) \left[ \mu_\star^2 \left[\eta^{1}(G^{-1}_W)\eta^{2} \right] + \mu_1^2 \psi_1 \left[\lambda^{1}  H_W \lambda^{2} \right] + 2 \mu_1^2\mu_\star^2 \psi_1 [\lambda^1 S_W \eta^2] \right] \\
    & \exp\left(-\frac{n}{2}\log \det(G_X) -\frac{n}{2}\log \det(G_W) -\frac{1}{2}  \psi_{1} \psi_{2} \lambda \sum (\eta^{\alpha }_h)^2 - \frac{1}{2} \eta^{\alpha}_h \eta^{\beta}_h \hat \lambda^{\alpha }_i \hat \lambda^{\beta}_i + i \sqrt P \hat \lambda^{\alpha }_i \lambda^{\alpha }_i  \right).
    \end{split}
\end{align*}

\subsubsection{Expression of the action and the prefactor}

To complete the computation we integrate with respect to $\hat \lambda^{\alpha }_i$, using again formulae \eqref{eq:integrals}:

\begin{align}
    \left[G_{\hat \lambda}\right]^{\alpha\beta}_{ii'} &= \delta^{ii'} \eta^{\alpha }_h \eta^{\beta}_h,\\
    \left[J_{\hat \lambda}\right]^{\alpha }_i &= i \sqrt P \lambda^{\alpha }_i.
\end{align}

This yields the final expression of the term:
\begin{align*}
\begin{split}
    \Psi_3^v = \frac{\psi_2}{D} &\int \left(\prod_{\alpha=1}^{2n} d\eta^\alpha \right) \left(\prod_{\alpha=1}^{2n} d\lambda^\alpha \right) \left(\mu_1^2 \psi_1\lambda^{1}_{i} \lambda^{2}_{i} + \mu_\star^2 \eta^{1}_{i} \eta^{2}_{i}\right) \left[ \mu_\star^2 \left[\eta^{1}(G^{-1}_W)\eta^{2} \right] + \mu_1^2 \psi_1 \left[\lambda^{1}  H_W \lambda^{2} \right] + 2 \mu_1^2\mu_\star^2 \psi_1 [\lambda^1 S_W \eta^2] \right] \\
    & \exp\left(-\frac{n}{2}\log \det(G_X) -\frac{n}{2}\log \det(G_W)-\frac{D}{2}\log \det(G_{\hat\lambda}) -\frac{1}{2}  \psi_{1} \psi_{2} \lambda \sum (\eta^{\alpha }_h)^2 -\frac{P}{2} \lambda^{\alpha }_{i} (G_{\hat\lambda}^{-1})^{\alpha\beta}_{ii'} \lambda^{\beta}_{i'} \right).
    \end{split}
\end{align*}

The above may be written as
\begin{align}
    \Psi_3^v = \int & \left(\prod d\eta \right) \left(\prod d\lambda \right)
    P_{\Psi_3^v} \left[ \eta, \lambda \right] \exp \left( {-\frac{D}{2} S^v\left[ \eta, \lambda \right]} \right),
\end{align}

with the \emph{prefactor} $P_{\Psi_3^v}$ and the \emph{action} $S^v$ defined as:
\begin{align*}
    P_{\Psi_3^v} \left[ \eta, \lambda \right]:=& \frac{\psi_2}{D} \left(\mu_1^2 \psi_1\lambda^{1}_{i} \lambda^{2}_{i} + \mu_\star^2 \eta^{1}_{i} \eta^{2}_{i}\right) \left[ \mu_\star^2 \left[\eta^{1}(G^{-1}_W)\eta^{2} \right] + \mu_1^2 \psi_1 \left[\lambda^{1}  H_W \lambda^{2} \right] + 2 \mu_1^2\mu_\star^2 \psi_1 [\lambda^1 S_W \eta^2] \right],\\
    S^v\left[ \eta, \lambda \right]:=& \psi_2 \log \det(G_X) + \psi_2 \log \det(G_W) + \log \det(G_{\hat\lambda}) + \frac{1}{D} \psi_{1} \psi_{2} \lambda \sum (\eta^{\alpha }_h)^2 + \frac{P}{D} \left( \lambda^{\alpha }_{i} (G_{\hat\lambda}^{-1})^{\alpha\beta}_{ii'} \lambda^{\beta}_{i'} \right).
\end{align*}

\subsubsection{Expression of the action and the prefactor in terms of order parameters}

Here we see that we have a factor $D\to\infty$ in the exponential part, which can be estimated using the saddle point method. Before doing so, we introduce the following order parameters using the Fourier representation of the delta-function:
\begin{align}
	1 &= \int dQ_{\alpha \beta}  d\hat Q_{\alpha \beta} e^{\hat Q_{\alpha \beta} (P Q_{\alpha \beta} - \eta^{\alpha }_h \eta^{\beta}_h)} ,\\
	1 &= \int dR_{\alpha \beta}  d\hat R_{\alpha \beta} e^{\hat R_{\alpha \beta} (D R_{\alpha \beta} - \lambda^{\alpha }_i \lambda^{\beta }_i)}.
\end{align}
This allows to rewrite the prefactor only in terms of $Q,R$: for example, $$\mu_1^2 \psi_1\lambda^{1}_{i} \lambda^{2}_{i} + \mu_\star^2 \eta^{1}_{i} \eta^{2}_{i} = \psi_1 D (\mu_1^2 R^{12} + \mu_\star^2 Q^{12}).$$ 
To do this, there are two key quantities we need to calculate: $\lambda G^{-1}_X \lambda$ and $\eta G^{-1}_W \eta$. To calculate both, we note that $G_X$ ang $G_W$ are both of the form $\mathbf{I}+\mathbf{X}$, therefore there inverse may be calculated using their series representation. The result is: 
\begin{align}
    \left[M_X\right]^{\alpha\beta} &:= \frac{1}{D} \lambda^{\alpha} G^{-1}_X\lambda^{\beta}
    = \left[ R (I+\mu_1^2 \psi_1 R)^{-1}\right]^{\alpha\beta} ,\\
    \left[M_W\right]^{\alpha\beta} &:= \frac{1}{P} \eta^{\alpha} (G_{W}^{-1}) \eta^{\beta} = \left[Q (I+\mu_\star^2\psi_1 AQ)^{-1}\right]^{\alpha\beta}.
\end{align}
Using the above, we deduce:
\begin{align}
    \lambda^1 H_W \lambda^2 &= D M_X^{12} + P \mu_1^2\mu_\star^2\psi_1 \left[M_X M_W M_X\right]^{12},\\
    \lambda^1 S_W \eta^2 &= P \left[M_X M_W \right]^{12}. 
\end{align}
The integrals over $\eta,\lambda$ become simple Gaussian integrals with covariance matrices given by $\hat Q, \hat R$, yielding:
\begin{align}
	1 &= \int dQ_{\alpha \beta}  d\hat Q_{\alpha \beta} e^{-\frac{\psi_1 D}{2}\left(\log\det\hat Q -2\mathrm{Tr} Q\hat Q\right)} ,\\
	1 &= \int dR_{\alpha \beta}  d\hat R_{\alpha \beta} e^{-\frac{D}{2}\left( \log\det\hat R -2\mathrm{Tr} R\hat R \right)}.
\end{align}

The next step is to take the saddle point with respect to the auxiliary variables $\hat Q$ and  $\hat R$ in order to eliminate them:
\begin{align}
    \frac{\partial S^v}{\partial\hat Q_{\alpha\beta}} = \psi_1 \left( \hat Q^{-1} - 2 Q \right) &= 0 \Rightarrow \hat Q = \frac{1}{2} Q^{-1},\\
    \frac{\partial S^v}{\partial\hat R_{\alpha\beta}} = \left( \hat R^{-1} - 2 R \right) &= 0 \Rightarrow \hat R = \frac{1}{2} R^{-1}.
\end{align}

One finally obtains that:
\begin{align}
    \Psi_3^v = \int & \left(\prod dQ \right) \left(\prod dR \right)
    P_{\Psi_3^v} \left[ Q,R \right] \exp \left( {-\frac{D}{2} S^v\left[ Q,R \right]} \right),
\end{align}
With:
\begin{align}
    P_{\Psi_3^v} \left[ Q, R \right] =& D \psi_1^2 \psi_2  (\mu_1^2 R^{12} + \mu_\star^2 Q^{12}) \left[\mu_\star^2  \left[M_W^v\right]^{12} + \mu_1^2 \left( M_X^{12} + \mu_1^2 \mu_\star^2 \psi_1^2 \left[M_X M_W^v M_X\right]^{12}\right) + 2 \mu_1^2\mu_\star^2 \psi_1 \left[M_X M_W^v\right]^{12} \right],\notag \\
    S^v\left[ Q, R \right] =& \psi_2 \log \det(G_X) + \psi_2 \log \det(G_W) +  \psi_{1}^2 \psi_{2} \lambda \mathrm{Tr} Q + \mathrm{Tr} \left( RQ^{-1} \right) +(1 - \psi_1) \log \det Q - \log \det R.
\label{eq:pre-act-q-r}
\end{align}

\subsubsection{Saddle point equations}
The aim is now to use the saddle point method in order to evaluate the integrals over the order parameters. Thus, one looks for $R$ and $Q$ solutions to the equations:
$$\frac{\partial S^v}{\partial Q_{\alpha\beta}}=0, \quad \frac{\partial S^v}{\partial R_{\alpha\beta}}=0\text{ }\text{ }\forall \alpha,\beta=1,\cdots,2n .$$
To solve the above, it is common to make a \emph{replica symmetric ansatz}. In this case, we assume that the solutions to the saddle points equations take the form:
\begin{align}
    Q=\left[\begin{array}{cccc}
    q & \tilde q&\cdots & \tilde q\\
    \tilde q& \ddots & \ddots & \vdots\\
    \vdots & \ddots & \ddots & \tilde q \\
    \tilde q & \cdots & \tilde q & q\end{array}\right], \quad
    R=\left[\begin{array}{cccc}
    r & \tilde r&\cdots & \tilde r\\
    \tilde r& \ddots & \ddots & \vdots\\
    \vdots & \ddots & \ddots & \tilde r \\
    \tilde r & \cdots & \tilde r & r\end{array}\right]
\end{align}
The action takes the following form:
\begin{align}
        S^v(q,r,\tilde q, \tilde r) &= 2n\left(S_0( q,r)+S_1^v( q,r,\tilde q,\tilde r)\right)\notag\\
        S_0(q,r)&=\lambda \psi_1^2 \psi_2   q+\psi_2 \log \left(\frac{\mu_\star^2  \psi_1q}{\mu_1^2  \psi_1r+1}+1\right)+\frac{r}{q}+(1-\psi_1) \log (q)+\psi_2 \log \left(\mu_1^2  \psi_1 r+1\right)-\log (r)\notag\\
        S_1^v( q,r,\tilde q,\tilde r)&=f^v(q,r) \tilde q + g^v(q,r)\tilde r\notag\\
        f^v(q,r)&=\lambda \psi_1^2 \psi_2+\frac{\mu_\star^2 \psi_1 \psi_2}{\mu_\star^2 \psi_1 q+\mu_1^2 \psi_1 r+1}+\frac{1-\psi_1}{q}-\frac{r}{q^2}\notag\\       g^v(q,r)&=-\frac{\mu_\star^2 \mu_1^2 \psi_1^2 \psi_2 q}{\left(\mu_1^2 \psi_1 r+1\right) \left(\mu_\star^2 \psi_1 q+\mu_1^2 \psi_1 r+1\right)}+\frac{\mu_1^2 \psi_1 \psi_2}{\mu_1^2 \psi_1 r+1}+\frac{1}{q}-\frac{1}{r}.
        \label{eq:action}
\end{align}

\subsubsection{Fluctuations around the saddle point}

We introduce the following notations:
\begin{align*}
    \left[\nabla_{\bs T} F(\bs T_\star)\right]_{\alpha \beta} &= \frac{\partial F}{\partial \bs T_{\alpha\beta}}|_{\bs T_\star},
    \\
    \left[H_{\bs T}F(\bs T_\star)\right]_{\alpha\beta,\gamma\delta}&=\at{\begin{bmatrix}
    \frac{\partial^2 F}{\partial Q_{\alpha \beta}\partial Q_{\gamma \delta}}&\frac{\partial^2 F}{\partial Q_{\alpha \beta}\partial R_{\gamma \delta}}\\
    \frac{\partial^2 F}{\partial R_{\alpha \beta}\partial Q_{\gamma \delta}}&\frac{\partial^2 F}{\partial R_{\alpha \beta}\partial R_{\gamma \delta}}
    \end{bmatrix}}{\bs T_\star},\\
    H\left[F\right]&=\at{\begin{bmatrix}
    \frac{\partial F}{\partial q\partial q} &\frac{\partial F}{\partial q\partial r}&\frac{\partial F}{\partial q\partial \tilde{q}}&\frac{\partial F}{\partial q\partial\tilde{r}} \\
    \frac{\partial F}{\partial q\partial r}&\frac{\partial F}{\partial r\partial r} &\frac{\partial F}{\partial r\partial \tilde{q}} &\frac{\partial F}{\partial r\partial\tilde{r}} \\
    \frac{\partial F}{\partial q\partial \tilde{q}} &\frac{\partial F}{\partial r\partial \tilde{q}} &\frac{\partial F}{\partial \tilde{q}\partial \tilde{q}} &\frac{\partial F}{\partial \tilde{q}\partial\tilde{r}} \\
    \frac{\partial F}{\partial q\partial \tilde{r}} &\frac{\partial F}{\partial r\partial \tilde{r}}&\frac{\partial F}{\partial \tilde{q}\partial \tilde{r}}&\frac{\partial F}{\partial \tilde{r}\partial\tilde{r}}
    \end{bmatrix}}{\substack{q=q^*\\r=r^*\\\tilde r=0 \\\tilde q=0\\\\}}.
\end{align*}

\paragraph{Proposition}
Let $q^\star$ and $r^\star$ be the solutions of the fixed point equation for the function $S_0:(q,r)\mapsto \mathbb R$ defined in \eqref{eq:action}:
\begin{align*}
    &\begin{cases}
    \frac{\partial S_0(q,r)}{\partial q}=0\\
    \frac{\partial S_0(q,r)}{\partial r}=0.
    \end{cases}\\
\end{align*}
Then we have that
\begin{equation}
    \Psi_3^v =\frac{1}{D}\operatorname{Tr}\left[H\left[S^v\right]^{-1}    H\left[P_{\Psi_{3}^v}\right]\right].\\
    \label{eq:hessian}
\end{equation}

\paragraph{Sketch of proof}

Solving the saddle point equations:
\begin{align*}
    &\begin{cases}
    \frac{\partial S^v(q,r,\tilde q,\tilde r)}{\partial q}=0\\
    \frac{\partial S^v(q,r,\tilde q,\tilde r)}{\partial q}=0\\
    \frac{\partial S^v(q,r,\tilde q,\tilde r)}{\partial \tilde q}=0\\
    \frac{\partial S^v(q,r,\tilde q,\tilde r)}{\partial \tilde r}=0
    \end{cases},
\end{align*}
one finds $\tilde q= \tilde r=0$, which is problematic because the prefactor vanishes: $P_{\Psi_3}\propto \mu_1^2\tilde q + \mu_\star^2 \tilde r$. 

Therefore we must go beyond the saddle point contribution to obtain a non zero result, i.e. we have to examine the quadratic fluctuations around the saddle point. To do so we preform a second-order expansion of the action \eqref{eq:pre-act-q-r} as a function of $Q$ and $R$:
\begin{align*}
    P_{\Psi_{3}^v}(\bs T) &\approx P_{\Psi_{3}^v}(\bs T_\star)+\bs (\bs T - \bs T_\star)^\top \nabla P_{\Psi_{3}^v}(\bs T_\star) + \frac{1}{2} (\bs T - \bs T_\star)^\top H_{\bs T}\left[P_{\Psi_{3}^v}(\bs T_\star)\right] (\bs T - \bs T_\star),\\
    S^v(\bs T)&\approx S_0(\bs T_\star)+\frac{1}{2}(\bs T-\bs T_\star)^\top H_{\bs T}\left[S^v(\bs T_\star)\right] (\bs T-\bs T_\star).
\end{align*}
Computing the second derivative of \eqref{eq:pre-act-q-r}, it is easy to show that:
\begin{align*}
    \left[H_{\bs T}\left[S^v(\bs T)\right]\right]_{\alpha\beta,\gamma\delta}&=\left[H_{\bs T}\left[S^v(\bs T)\right]\right]_{\alpha\beta} \left(\delta_{\alpha\gamma}\delta_{\beta\delta}+\delta_{\alpha\delta}\delta_{\beta\gamma}\right),\\
    \left[H_{\bs T}\left[P_{\Psi_3^v}(\bs T)\right]\right]_{\alpha\beta,\gamma\delta}&=\left[H_{\bs T}\left[P_{\Psi_3^v}(\bs T)\right]\right]_{\alpha\beta} \left(\delta_{\alpha\gamma}\delta_{\beta\delta}+\delta_{\alpha\delta}\delta_{\beta\gamma}\right),
\end{align*}
where
\begin{align*}
    H_{\bs T}\left[F\right]_{\alpha \beta} &= \begin{bmatrix}
    \frac{\partial^2 F}{\partial Q_{\alpha \beta}\partial Q_{\alpha \beta}}&\frac{\partial^2 F}{\partial Q_{\alpha \beta}\partial R_{\alpha \beta}}\\
    \frac{\partial^2 F}{\partial R_{\alpha \beta}\partial Q_{\alpha \beta}}&\frac{\partial^2 F}{\partial R_{\alpha \beta}\partial R_{\alpha \beta}}\end{bmatrix}\\
    &= \frac{1}{2n}\delta_{\alpha \beta}\begin{bmatrix}
    \frac{\partial^2 F}{\partial q\partial q}&\frac{\partial^2 F}{\partial q\partial r}\\
    \frac{\partial^2 F}{\partial r\partial q}&\frac{\partial^2 F}{\partial r\partial r}\end{bmatrix} + \frac{2}{2n(2n-1)}(1-\delta_{\alpha\beta})\begin{bmatrix}
    \frac{\partial^2 F}{\partial \tilde q\partial \tilde q}&\frac{\partial^2 F}{\partial \tilde q\partial\tilde r}\\
    \frac{\partial^2 F}{\partial\tilde r\partial \tilde q}&\frac{\partial^2 F}{\partial\tilde r\partial\tilde r}\end{bmatrix}.
\end{align*}
Hence, 
\begin{align*}
    \Psi_3^v &= \lim_{n\to 0}\int d\bs T P_{\Psi_3^v}(\bs T) \exp^{-\frac{D}{2} S^v(\bs T)}, \\
    &= \lim_{n\to 0}\underbrace{P_{\Psi_3^v}(\bs T_\star)}_\text{0} + \nabla P_{\Psi_3^v}(\bs T_\star)_{\alpha\beta} \underbrace{\int d\bs T (\bs T - \bs T_\star)_{\alpha\beta} \exp\left(-\frac{D}{2} S^v(\bs T)\right)}_\text{0} \\&\quad + \frac{1}{2}H_{\bs T}\left[P_{\Psi_3^v}(\bs T_\star)\right]_{\alpha\beta} \int d\bs T (\bs T - \bs T_\star)_{\alpha \beta}^2 \exp\left(-\frac{D}{2} S^v(\bs T)\right)\\
    &= \lim_{n\to 0}\frac{1}{2}H_{\bs T} \left[P_{\Psi_3^v}(\bs T_\star)\right]_{\alpha\beta} e^{-\frac{D}{2}S^v(\bs T_\star)}\int d\bs T (\bs T - \bs T_\star)_{\alpha \beta}^2 e^{-\frac{D}{2}\mathop{\sum}_{\alpha\beta}\bs (\bs T-\bs T_\star)_{\alpha\beta}^2 H_{\bs T}\left[S^v(\bs T_\star)\right]_{\alpha\beta}}  \\
    &= \lim_{n\to 0} \frac{1}{2}\frac{(2\pi)^{n}}{\det H_{\bs T}\left[S^v(\bs T_\star)\right]} e^{-\frac{D}{2}S^v(\bs T_\star)} \frac{2}{D}H_{\bs T}\left[P_{\Psi_3^v}(\bs T_\star)\right]_{\alpha\beta} H_{\bs T}\left[S^v(\bs T_\star)\right]_{\alpha\beta}^{-1}\\
    &=\frac{1}{D}\operatorname{Tr}\left[H\left[S^v\right]^{-1}    H\left[P_{\Psi_{3}^v}\right]\right]
\end{align*}
In the last step, we used the fact that:
\begin{align*}
    \lim_{n\to 0} e^{-\frac{D}{2}S^v(\bs T_\star)}&=\lim_{n\to 0} e^{-nD S_0(q_\star, r_\star)}=1,\\
    \lim_{n\to 0} \det H_{\bs T}\left[S^v(\bs T_\star)\right] &= 1.
\end{align*}
The last equality follows from the fact that for a matrix of size $n\times n$ of the form $M_{\alpha\beta} = a\delta_{\alpha\beta} + b(1-\delta_{\alpha\beta})$, we have $$\det M = (a-b)^n \left(1+ \frac{nb}{a-b}\right) \xrightarrow[n \to 0]{} 1.$$

\subsubsection{Expression of the vanilla terms}
Using the above procedure, one can compute the terms $\Psi_1, \Psi_2^v, \Psi_3^v$ of \eqref{eq:ensemble-error-app}: for each of these terms, the action is the same as in \eqref{eq:pre-act-q-r}, and the prefactors can be obtained as:
\begin{align*}
&P_{\Psi_1}\left[Q,R\right]=\mu_1^2 \psi_1 \psi_2\left(M_X^{11}+\mu_1^2 \mu_\star^2 \psi_1^2 \left(M_XM_WM_X\right)^{11}+\mu_\star^2 \mu_1 \psi_1 \left(M_XM_W\right)^{11}\right),\\
&P_{\Psi_2^v}[Q,R]=D\psi_1^2\psi_2\left(\mu_1^2R^{12}+\mu_\star^2Q^{12}\right)\left(\mu_1^2 \psi_2 P_{XX}-2\mu_1^2\mu_\star^2\psi_1 \psi_2 P_{XW}+\mu_\star^2P_{WW}  \right),\\
&P_{\Psi_3^v} \left[ Q, R \right] = D \psi_1^2 \psi_2  (\mu_1^2 R^{12} + \mu_\star^2 Q^{12}) \left[\mu_\star^2  \left[M_W^v\right]^{12} + \mu_1^2 \left( M_X^{12} + \mu_1^2 \mu_\star^2 \psi_1^2 \left[M_X M_W^v M_X\right]^{12}\right) + 2 \mu_1^2\mu_\star^2 \psi_1 \left[M_X M_W^v\right]^{12} \right],\\
&P_{XX}=N_X^{12}+\frac{1}{\psi_2}M_X^{12}+2 \left(\mu_1\mu_\star\psi_1\right)^2\left[N_XM_WM_X\right]^{12}+\frac{\left(\mu_1\mu_\star\psi_1\right)^2}{\psi_2}[M_XM_WM_X]^{12}+\left(\mu_1\mu_\star\psi_1\right)^4[M_XM_WN_XM_WM_X]^{12},\\
&P_{XW}=\left[N_XM_W\right]^{12}+\frac{1}{\psi_2}\left[M_XM_W\right]^{12}+\left(\mu_1\mu_\star\psi_1\right)^2\left[M_XM_WN_XM_W\right]^{12},\\
&P_{WW}=M_W^{12}+\psi_2 \left(\mu_1\mu_\star\psi_1\right)^2 \left[M_WN_XM_W\right]^{12}.
\end{align*}
Where a new term appears:
\begin{align}
        [N_X]^{\alpha\beta}&=\left[ R (I+\mu_1^2 \psi_1 R)^{-2}\right]^{\alpha\beta}.
\end{align}

\subsection{Computation of the \emph{ensembling} terms}

\subsubsection{Expression of the action and the prefactor}
In the \emph{ensembling} terms, the two inverse matrices are different, hence one has to introduce two distinct replica variables. We distinguish them by the use of an extra index $a\in\lbrace1,2\rbrace$, denoted in brackets in order not to be confused with the replica indices~$\alpha$.
\begin{equation}
\left[M^{(1)}\right]^{-1}_{ij} \left[M^{(2)}\right]^{-1}_{kl} = \lim_{n\to 0} \int \left(\prod_{\alpha=1}^{n}\prod_{a=1}^{2} d\eta^{\alpha(a)}\right) \eta^{1(1)}_i \eta^{1(1)}_j \eta^{1(2)}_k \eta^{1(2)}_l \exp\left(-\frac{1}{2}\sum_{(a)} \eta^{\alpha(a)} M^{(a)}_{ij} \eta^{\alpha(a)}\right).
\label{eq:replica_mixed}
\end{equation}

Calculations of the Gaussian integrals follow through in a very similar way as for the \emph{vanilla} terms. The matrices appearing in the process are:

\begin{align}
    &(G^e_X)_{ii'} = \delta_{i i'} + \frac{\mu_1^2 \psi_1}{D} \sum_{(a)\alpha} \lambda^{\alpha (a)}_i \lambda^{\alpha (a)}_{i'},\\
    &(J^e_X)_{i} = \frac{\mu_1 \mu_\star \sqrt{\psi_1}}{D^{2}} \sum_{\alpha a} \lambda^{\alpha (a)}_i \eta^{\alpha (a)}_{h} W^{(a)}_{h},\\
    &(G^e_W)^{(ab)}_{hh'} = \delta^{ab}_{hh'} +\frac{\mu_\star^2}{D} \sum_{\alpha\beta} \eta^{\alpha (a)}_h A_e^{\alpha \beta, ab} \eta^{\beta (b)}_{h'} ,\\
    &(J^e_W)_{h} = 0\\
    &(G^e_\Theta)^{(ab)}_{hh', ii'} = \delta^{ab}_{hh', ii'}\\
    &(J^e_\Theta)^{(a)}_{hi} = - i \sum_\alpha \hat \lambda^{\alpha (a)}_i \eta^{\alpha (a)}_h,\\
    &(G^e_{\hat \lambda})^{\alpha\beta, (ab)}_{ii'} = 2\delta^{ab} \delta^{ii'} \eta^{\alpha (a)}_h \eta^{\beta (b)}_h,\\
    &(J^e_{\hat \lambda})^{\alpha (a)}_i = i \sqrt P \lambda^{\alpha (a)}_i.\\
\end{align}
with
\begin{align}
    &A_e^{\alpha \beta, (ab)}=\delta_{ab}\delta_{\alpha\beta} -\mu_1^2 \psi_1 \frac{1}{D} \sum_{i,j} \lambda_i ^{\alpha (a)}[G_X^{e-1}]_{ij} \lambda_j ^{\beta (b)},\\
    &\left[H^e_W\right]_{ii'} = \left[G^{e-1}_X\right]_{i i'} + \sum_{\alpha\beta, ab} \left[ G^{e-1}_X \lambda^{\alpha (a)}\right]_i \left[ G^{e-1}_X \lambda^{\beta (b)}\right]_{i'} \left[\eta^{\alpha (a)} \left[G^{e-1}_W\right]^{(ab)} \eta^{\beta (b)}\right],\\
    &[S^e_W]_{ih} =\frac{1}{D} \sum_{\alpha, a}\left[ G^{e-1}_X\lambda^{\alpha(a)}\right]_i \left[ G_W^{e-1}\eta^{\alpha(a)}\right]_h .
\end{align}
Starting with the computation of $\Psi_3^e$ in order to illustrate the method used, the prefactor $P_{\Psi_3^e}$ and the action are $S^e$ are given by:
\begin{align}
    P_{\Psi^e_3} \left[ \eta, \lambda \right]:=& \frac{\mu_1^2 \psi_1\psi_2}{D} \lambda^{1(1)}_{i} \lambda^{1(2)}_{i} \left[ \mu_\star^2 \left[\eta^{1(1)}(G^{e-1}_W)^{(12)}\eta^{1(2)} \right] + \mu_1^2 \psi_1 \left[\lambda^{1(1)}  H^e_W \lambda^{1(2)} \right] + 2 \mu_1^2\mu_\star^2 \psi_1 [\lambda^{1(1)} S^e_W \eta^{1(2)}]\right] ,\\
    S^e\left[ \eta, \lambda \right]:=& \psi_2 \log \det(G^e_X) + \psi_2 \log \det(G^e_W) + \log \det(G^e_{\hat\lambda}) + \frac{1}{D} \psi_{1} \psi_{2} \lambda \sum (\eta^{\alpha (a)}_h)^2 + \frac{1}{2D} \left( \lambda^{\alpha (a)}_{i} (G_{\hat\lambda}^{e-1})^{\alpha\beta, ab}_{ii'} \lambda^{\beta (b)}_{i'} \right).
\end{align}

\subsubsection{Expression of the action and the prefactor in terms of order parameters}
This time, because of the two different systems, the order parameters carry an additional index $a$, which turns them into $2\times2$ block matrices:
\begin{align}
	1 &= \int dQ^{(ab)}_{\alpha \beta}  d\hat Q^{(ab)}_{\alpha \beta} e^{\hat Q^{(ab)}_{\alpha \beta} (P Q^{(ab)}_{\alpha \beta} - \eta^{\alpha (a)}_h \eta^{\beta (b)}_h)}, \\
	1 &= \int dR^{(ab)}_{\alpha \beta}  d\hat R^{(ab)}_{\alpha \beta} e^{\hat R^{(ab)}_{\alpha \beta} (d R^{(ab)}_{\alpha \beta} - \lambda^{\alpha (a)}_i \lambda^{\beta (b)}_i)}.
\end{align}
The systems being decoupled, we make the following ansatz for the order parameters:
\begin{align}
Q=\left[
\begin{array}{c|c}
\begin{array}{ccc}
q&\cdots&0\\
\vdots &\ddots & \vdots\\
0&\cdots&q\\
\end{array} & \begin{array}{ccc}
\tilde q&\cdots&0\\
\vdots &\ddots & \vdots\\
0&\cdots&\tilde q\\
\end{array} \\ \hline
\begin{array}{ccc}
\tilde q&\cdots&0\\
\vdots &\ddots & \vdots\\
0&\cdots&\tilde q\\
\end{array} & \begin{array}{ccc}
q&\cdots&0\\
\vdots &\ddots & \vdots\\
0&\cdots&q\\
\end{array}
\end{array}\right], \quad
R=\left[
\begin{array}{c|c}
\begin{array}{ccc}
r&\cdots&0\\
\vdots &\ddots & \vdots\\
0&\cdots&r\\
\end{array} & \begin{array}{ccc}
\tilde r&\cdots&0\\
\vdots &\ddots & \vdots\\
0&\cdots&\tilde r\\
\end{array} \\ \hline
\begin{array}{ccc}
\tilde r&\cdots&0\\
\vdots &\ddots & \vdots\\
0&\cdots&\tilde r\\
\end{array} & \begin{array}{ccc}
r&\cdots&0\\
\vdots &\ddots & \vdots\\
0&\cdots&r\\
\end{array}
\end{array}\right].
\label{eq:orderparammixed}
\end{align}

In virtue of the simple structure of the above matrices, the replica indices $\alpha$ trivialize and we may replace the matrices $Q$ and $R$ by the $2\times 2$ matrices:
$$Q=\begin{bmatrix}
   q&\tilde q\\\tilde q&q\end{bmatrix},
   \text{  }\text{  }
   R=\begin{bmatrix}
   r&\tilde r\\\tilde r&r
\end{bmatrix}.$$\\
Define:
\begin{align}
    \left[M^e_X\right]_{(ab)} &\equiv \frac{1}{D} \lambda_i^{(a)} \left[G_X^{e-1}\right]_{ij}\lambda_j^{(b)} =  \left[ R (I+\mu_1^2 \psi_1 R)^{-1}\right]_{(ab)}, \\
    \left[M^e_W\right]_{(ab)} &\equiv \frac{1}{P} \eta^{(a)} \left[G_{W}^{e-1}\right]^{(ab)} \eta^{(b)} = \left[Q (I+\psi_1 A^eQ)^{-1}\right]_{(ab)},
\end{align}
where products are now over $2\times 2$ matrices. Then, one has:
\begin{align*}
    P_{\Psi_3^e}\left[ Q, R \right]:=& D \mu_1^2 \psi_1^2 \psi_2 R^{(12)} \left[\mu_\star^2  M_W^{e(12)} + \mu_1^2 \left( M_X^{e(12)} + \mu_1^2 \mu_\star^2 \psi_1^2 \left[M^e_X M^e_W M^e_X\right]^{(12)}\right) + 2\mu_1^2\mu_\star^2 \psi_1 \left[M^e_X M^e_W\right]^{(12)} \right],\\
    \begin{split}
        S^e\left[ Q, R \right]:=&\psi_2 \log \det(G^e_X) + \psi_2 \log \det(G^e_W) + \sum_a \left[ (\psi_{1}^2 \psi_{2} \lambda) \mathrm{Tr} Q^{(a)(a)} + \mathrm{Tr} \left( R^{(a)(a)} (Q^{-1})^{(a)(a)}\right) + \log \det Q^{(a)(a)}  \right]\\&- \psi_1 \log \det Q - \log \det R.
    \end{split}
\end{align*}
Where we have:
\begin{align}
    M^e_X&=\frac{1}{(1 + \mu_1^2 \psi_1 r)^2 - (\mu_1^2 \psi_1 \tilde r)^2}\begin{bmatrix}
    r + \mu_1^2 \psi_1 (r^2-\tilde r^2)  & \tilde r\\
    \tilde r & r + \mu_1^2 \psi_1 (r^2-\tilde r^2)
    \end{bmatrix},\\
    A^e_{(a b)}&=\delta_{(ab)} -\mu^2_1 \psi_1 \left[M^e_X\right]_{(ab)} ,\\
    \left[M^e_W\right]_{(ab)} &= q \delta_{(ab)} - \mu_\star^2 \psi_1 \left[A^e(I+ \mu_\star^2\psi_1 q A^e)^{-1}\right]_{(ab)},\\
    \det(G^e_X)&=\det(\delta_{(ab)}+ \psi_1 \mu_1^2 R_{(a b)}),\\
    \det(G^e_W)&=\det(\delta_{a b}+ \psi_1 q A^e_{(ab)}).
\end{align}
Finally, we are left with:
\begin{align*}
    S^e(q,r,\tilde q, \tilde r) =&n \left(S_0(q,r) + S_1^e(q,r,\tilde q, \tilde r)\right),\\
S_1^e(q,r,\tilde{q},\tilde r)=& \Tilde{r}^2 f^e(q,r) + \Tilde{q}^2 g^e(q,r),\\
 f^e(q,r)=&\frac{2 r \mu_1^{2} \psi_1 \left( 1+q \mu_\star^2\psi_1\right)+\left(1+q \mu_\star^2\psi_1\right)^{2}-r^{2} \mu_1^{4} \psi_1^{2}(-1+\psi_2)}{2r^{2}\left(1+r \mu_1^{2} \psi_1+q \mu_\star^2\psi_1\right)^{2}},\\
 g^e(q,r)=&\frac{\psi_1}{2 q^2}.
\end{align*}
Where $S_0$ was defined in \eqref{eq:action}.

\subsubsection{Expression of the ensembling terms}

Evaluating the fluctuations around the saddle point follows through in the same way as for the vanilla terms, with the following expressions of the prefactors:
\begin{align}
    P_{\Psi_2^e}\left[ Q, R \right] &= D \psi_1^2 \psi_2  \mu_1^2 \tilde r\left[\mu_1^2 P^e_{XX}-2 \mu_1^2 \mu_\star^2 \psi_1 P^e_{WX} + \mu_\star^2 P^e_{WW}\right],\\
    P_{\Psi_3^e}\left[ Q, R \right] &= D \mu_1^2 \psi_1^2 \psi_2 R^{(12)} \left[\mu_\star^2  M_W^{e(12)} + \mu_1^2 \left( M_X^{e(12)} + \mu_1^2 \mu_\star^2 \psi_1^2 \left[M^e_X M^e_W M^e_X\right]^{(12)}\right) + 2\mu_1^2\mu_\star^2 \psi_1 \left[M^e_X M^e_W\right]^{(12)} \right],\\
    \begin{split}
    P^e_{XX} &= \psi_2 N_X^{e12} +M_X^{e12}+ 2 \psi_2(\mu_1 \mu_\star\psi_1)^2 \left[M^e_X N^e_X M_W^e\right]^{12} +(\mu_1 \mu_\star\psi_1)^2\left[M^e_X M_W^e M^e_X\right]^{12}\\
    &+ \psi_2(\mu_1 \mu_\star\psi_1)^4 \left[M^e_X M^e_W N^e_X M^e_W M^e_X\right]^{12},
    \end{split}\\
    P^e_{WX} &=\psi_2 \left[N^e_X M_W^e\right]^{12} + \left[M^e_X M_W^e\right]^{12}+\psi_2 (\mu_1 \mu_\star\psi_1)^2 \left[M^e_X M_W^e N^e_X M_W^e\right]^{12} ,\\
P^e_{WW} &= [M_W^e]^{12}+ \psi_2(\mu_1 \mu_\star\psi_1)^2 \left[M_W^e N^e_X M^e_W\right]^{12}.
\end{align}

\subsection{Computation of the \emph{divide and conquer} term}

Here, we are interested in computing the term $\Psi_2^d$. This term differs from the previous ones in that there are now two independent data matrices $\boldsymbol{X}^{(1)}$ and $\boldsymbol{X}^{(2)}$. The calculations for the action and the prefactor are very similar to calculations performed for the \emph{ensembling} terms $\Psi_2^e, \Psi_3^e$, with the addition that $\bs X$ now also carries an index $(a)$. \\
Firstly let us write $\Psi_2^d$ as a trace over random matrices:
\begin{equation}
    \Psi_2^d=\frac{\mu_1^2}{d^2}\operatorname{Tr}\left[\boldsymbol{X}^{(1)}\boldsymbol{Z}^{(1)} \boldsymbol{B}^{(1)}{}^{-1} \boldsymbol{\Theta}^{(1)}\boldsymbol{\Theta}^{(2)}\boldsymbol{B}^{(2)}{}^{-1}\boldsymbol{Z}^{(2)}\boldsymbol{X}^{(2)} \right].
\end{equation}
Calculations follow through in the same way as in the previous sections. Using the replica formula \eqref{eq:replica_mixed}, and performing the integrals over the Gaussian variables the following quantities appear:
\begin{align}
    &(G^d_X)^{(ab)}_{ii'} = \delta_{i i'} + \frac{\mu_1^2 \psi_1}{D} \sum_{\alpha} \lambda^{\alpha (a)}_i \lambda^{\alpha (a)}_{i'},\\
    &(J^{d}_X)^{(a)}_{i} = \frac{\mu_1 \mu_\star \sqrt{\psi_1}}{D^{2}} \sum_{\alpha} \lambda^{\alpha (a)}_i \eta^{\alpha (a)}_{h} W^{(a)}_{h},\\
    &(G^d_W)^{(ab)}_{hh'} = \delta^{ab}_{hh'} +\frac{\mu_\star^2}{D} \sum_{\alpha\beta} \eta^{\alpha (a)}_h A^{\alpha \beta, ab} \eta^{\beta (b)}_{h'},\\
    &(J^d_W)_{h} = 0,\\
    &(G^d_\Theta)^{(ab)}_{hh', ii'} = \delta^{ab}_{hh', ii'},\\
    &(J^d_\Theta)^{(a)}_{hi} = - i \sum_\alpha \hat \lambda^{\alpha (a)}_i \eta^{\alpha (a)}_h,\\
    &(G^d_{\hat \lambda})^{\alpha\beta, (ab)}_{ii'} = 2\delta^{ab} \delta^{ii'} \eta^{\alpha (a)}_h \eta^{\beta (b)}_h,\\
    &(J^d_{\hat \lambda})^{\alpha (a)}_i = i \sqrt P \lambda^{\alpha (a)}_i.\\
\end{align}
The saddle point ansatz for $Q$ and $R$ is the same as the one for the ensembling terms (see \eqref{eq:orderparammixed}). The procedure to evaluate $\Psi_2^d$ is also the same as the one for $\Psi_2^e$ except the Hessian is taken with respect to $S^d$. The final result is given below.
\begin{align*}
P_{\Psi_2^d}\left[Q,R\right] =& D \mu_1^2  \psi_1 \psi_2^2  \tilde{r} \left[ \psi_1 \mu_1^2 P_{XX} + 2 \mu_\star^2 \mu_1^2 \psi_1^2 P_{WX} + \mu_\star^2 \psi_1 P_{WW} \right],\\
    P_{XX} =& \left( N_X^{11} + 2(\mu_1 \mu_\star\psi_1)^2 \left[N_X M_W M_X\right]^{11} + (\mu_1 \mu_\star\psi_1)^4 \left[M_X M_W N_X M_W M_X\right]^{11}\right),\\
    P_{WX} =&  \left[N_X M_W\right]^{11} + (\mu_1 \mu_\star\psi_1)^2 \left[M_X M_W N_X M_W\right]^{11},  \\
    P_{WW} =&(\mu_1 \mu_\star\psi_1)^2 \left[M_W N_X M_W\right]^{11}, \\
S^d[q,r,\Tilde{q},\Tilde{r}]=&n \left(S_0(q,r)+  S_1^d(q,r,\tilde{q},\tilde{r})\right),\\
S_1^d(q,r,\tilde{q},\tilde{r})=&\frac{\tilde r^2}{r^2}+\frac{\psi_1 \tilde q^2}{q^2}.
\end{align*}
With:
\begin{align*}
    M_X&= \frac{r}{1 + \mu_1^2 \psi_1 r}\ ,\\
    A&=1 -\mu^2_1 \psi_1 M_X\ ,\\
    M_W&= \frac{q}{1+\mu_\star^2 \psi_1qA}\ ,\\
    N_X&=\frac{\tilde r}{(1+\mu_1^2 \psi_1 r)^2}\ .\\
\end{align*}


\end{document}